%% file: main.tex
\documentclass[runningheads]{llncs}

\usepackage{eccv}

\usepackage{eccvabbrv}

\usepackage{graphicx}
\usepackage{booktabs}
\usepackage[colorinlistoftodos,prependcaption,textsize=tiny]{todonotes}
\usepackage{wrapfig}

\usepackage[accsupp]{axessibility}  

\usepackage{multirow}
\usepackage{multicol}
\usepackage{color, colortbl}
\definecolor{LightCyan}{rgb}{0.88,1,1}
\definecolor{LightGreen}{rgb}{0.56, 0.93, 0.56}

\newcommand{\SMILE}{SMILe}

\newcommand{\supcon}{SupCon}

\usepackage{hyperref}

\usepackage{orcidlink}

\begin{document}

\title{\SMILE: Leveraging Submodular Mutual Information For Robust Few-Shot Object Detection} 

\titlerunning{SMILe: Submodular Mutual Information Learner}

\author{Anay Majee\inst{1}\orcidlink{0000-0003-0189-8310} \and
Ryan Sharp\inst{2}\orcidlink{0009-0008-4871-8085}\thanks{Work done as a Graduate student at The University of Texas at Dallas.} \and
Rishabh Iyer\inst{1}\orcidlink{0000-0001-9851-463X}}

\authorrunning{Majee et al.}

\institute{
    The University of Texas at Dallas, TX, USA \\
    \email{\{anay.majee, rishabh.iyer\}@utdallas.edu} \and
    IGS Energy, OH, USA \\
    \email{rysharp@igsenergy.com}
}

\maketitle

\begin{abstract}
  \input{sections/01-abstract}
\end{abstract}

\section{Introduction}
\input{sections/02-introduction}

\section{Related Work}
\label{sec:rel_work}
\input{sections/03-related_work}

\section{Method}
\label{sec:method}
\input{sections/04-method}

\section{Experiments}
\label{sec:experiments}
\input{sections/05-experiment}

\section{Conclusion}
\input{sections/06-conclusion}

\section*{Acknowledgements}
We gratefully thank anonymous reviewers for their valuable comments. This work is supported by the National Science Foundation under Grant Numbers IIS-2106937, a gift from Google Research, an Amazon Research Award, and the Adobe Data Science Research award. Any opinions, findings, and conclusions or recommendations expressed in this material are those of the authors and do not necessarily reflect the views of the National Science Foundation, Google or Adobe.

%
%
\bibliographystyle{splncs04}
\bibliography{references}

\section*{Appendix}
\input{sections/07-supplementary}

\end{document}

%% file: sections/01-abstract.tex
Confusion and forgetting of object classes have been challenges of prime interest in Few-Shot Object Detection (FSOD).
To overcome these pitfalls in metric learning based FSOD techniques, we introduce a novel \textbf{S}ubmodular \textbf{M}utual \textbf{I}nformation \textbf{L}earning (\textbf{\SMILE}~\footnote{Project page:~\url{https://anaymajee.me/assets/project_pages/smile.html}.}) framework for loss functions which adopts combinatorial mutual information functions as learning objectives to enforce learning of well-separated feature clusters between the base and novel classes. Additionally, the joint objective in \SMILE\ minimizes the total submodular information contained in a class leading to discriminative feature clusters. 
The combined effect of this joint objective demonstrates significant improvements in class confusion and forgetting in FSOD.
Further we show that \SMILE\ generalizes to several existing approaches in FSOD, improving their performance, agnostic of the backbone architecture.
Experiments on popular FSOD benchmarks, PASCAL-VOC and MS-COCO show that our approach generalizes to State-of-the-Art (SoTA) approaches \textit{improving their novel class performance by up to 5.7\% (3.3 $mAP$ points) and 5.4\% (2.6 $mAP$ points) on the 10-shot setting of VOC (split 3) and 30-shot setting of COCO datasets} respectively. 
Our experiments also demonstrate better retention of base class performance and \textit{up to $2\times$ faster convergence} over existing approaches agnostic of the underlying architecture.\looseness-1

%% file: sections/02-introduction.tex
Recent advances in Deep Neural networks (DNNs) have enabled models to learn discriminative feature representations from \textit{large-scale} image benchmarks. 
Unfortunately, these architectures fail to adapt to few-shot settings tasked to recognize novel objects over existing ones with few examples, closely resembling human-like perception.
Although recent research has shown significant promise in few-shot image recognition \cite{maml, nonforget, protonet, relation-net, matching-net}, Few-Shot Object Detection (FSOD) remains a challenge with recent works \cite{fsdet, metadet, digeo, imted} highlighting two major challenges - \textit{Class Confusion} and \textit{Catastrophic Forgetting}.
\textit{Class confusion}, as highlighted in \cite{majee2021fewshot} manifests itself through mis-prediction of instances belonging to a newly learnt (novel) class, as one or more instances of the already learnt (base) classes.
Authors in \cite{agcm, metadet} attribute this to the sharing of visual information between classes resulting in \textbf{increased inter-class bias} due to overlapping feature clusters as shown in \cref{fig:comb_optim_functions}(a).
\textit{Catastrophic forgetting} refers to the gradual degradation in the performance of already learnt classes in the quest to learn the novel ones, as shown in \cref{fig:forget_converence}(a), seldom overfitting to rare classes \cite{metadet, majee2021fewshot}.
Further, \textbf{large feature diversity (intra-class variance) among base classes} lead to formation of non-discriminative feature clusters as shown in \cref{fig:comb_optim_functions}(b), aggravating the existing inter-class bias in the feature space.
Unlike existing approaches (refer \cref{sec:rel_work}) which target either confusion or forgetting, our paper presents a unified approach to tackle both these challenges in FSOD.\looseness-1
Although, recent approaches \cite{fscontrastive, digeo, agcm} attempt to tackle these challenges through contrastive learning strategies, such approaches have been limited by their capability to overcome either inter-class bias or intra-class variance \cite{score, metadet} and poor generalization to longtail settings~\cite{score} (FSOD being a extreme case).\looseness-1 

\input{figures/comb_optim_functions}

In this paper, we introduce a combinatorial viewpoint in FSOD considering each object class $i \in [1, C]$ in the dataset $\mathcal{T}$ as a set $A_i$ of samples, where $\mathcal{T} = \{A_1, \cdots A_C\}$, facilitating the application of combinatorial functions as learning objectives. 
We aim to overcome the aforementioned challenges through representation learning in the low-data regime by adopting this formulation through the \textbf{\SMILE}: \textbf{S}ubmodular \textbf{M}utual \textbf{I}nformation \textbf{Le}arning framework, wherein we introduce novel, set-based combinatorial objective functions for FSOD as shown in \cref{fig:smile_overview}.
\textbf{\SMILE\ introduces a joint objective formulation $L_{comb}$ (\cref{eq:info_score_objective}) based on two popular flavors of submodular information functions - Submodular Mutual Information \cite{prism} (SMI) and Total Submodular Information \cite{fujishige}} targeting the root causes of confusion and forgetting in FSOD.
At first, \SMILE\ is the first to introduce pairwise SMI functions $I_f$ in representation learning which model the common (overlapping) information between two sets. 
Minimizing $I_f$ through the joint objective $L_{comb}$ reduces feature overlap between \textit{base} and \textit{novel} classes alleviating inter-class bias in the model towards abundantly sampled classes as shown in \cref{fig:comb_optim_functions}(b).
We extend this property of SMI functions to the \textit{novel} classes minimizing the inter-cluster overlap between few-shot classes, promoting learning of discriminative features from just few samples.
Secondly, \SMILE\ preserves the diversity within each class by minimizing the total submodular information contained within each set as shown in \cref{fig:comb_optim_functions}(c), minimizing the impact of forgetting.
This formulation closely follows the observation in \cite{score} which models cooperation \cite{submod_cooperation} between instances in a set by minimizing a submodular function over a set, to preserve representative features.
The unified objective $L_{comb}$ introduced in \SMILE\ models both these necessary properties through a weighted sum of two distinct objectives $L_{comb}^{inter}$ and $L_{comb}^{intra}$ as shown in \cref{fig:smile_overview} balancing the tradeoff between inter-cluster separation and intra-cluster compactness respectively. 
This allows us to introduce a family of loss functions which inherently eliminates confusion and forgetting as shown in \cref{tab:abl_smi}. 
We conduct our experiments on two popular FSOD benchmarks, PASCAL-VOC \cite{pascalvoc} and MS-COCO \cite{mscoco} for several few-shot settings and demonstrate the following contributions of \SMILE:

\input{figures/forgetting_convergence}

\begin{itemize}
\item \textbf{\SMILE\ introduces a novel set-based combinatorial viewpoint in FSOD} by applying combinatorial Mutual Information based objective to discriminate between base and novel classes, in conjunction with submodular total information to minimize intra-class variance as the objective function.\looseness-1
\item \SMILE\ \textbf{generalizes to existing approaches in FSOD, irrespective of the underlying architecture} demonstrating up to 5.7\% improvement in novel class performance (\cref{tab:voc}) over the baseline FSOD approach.\looseness-1
\item \SMILE\ demonstrates up to \textbf{2$\times$ faster convergence} (\cref{fig:forget_converence}(b)) over existing SoTA approaches resulting in faster generalization to unknown object classes.\looseness-1
\item Finally, \SMILE\ demonstrates up to \textbf{11\% and 3.5\% reduction in class confusion and catastrophic forgetting} while achieving SoTA performance on popular FSOD benchmarks like PASCAL-VOC (by 5.7\% on split 2, 10-shot setting) and MS-COCO (5.4\% on 30-shot setting).
\end{itemize}

%% file: figures/comb_optim_functions.tex
\begin{figure*}[t]
        \centering
        \includegraphics[width=0.9\textwidth]{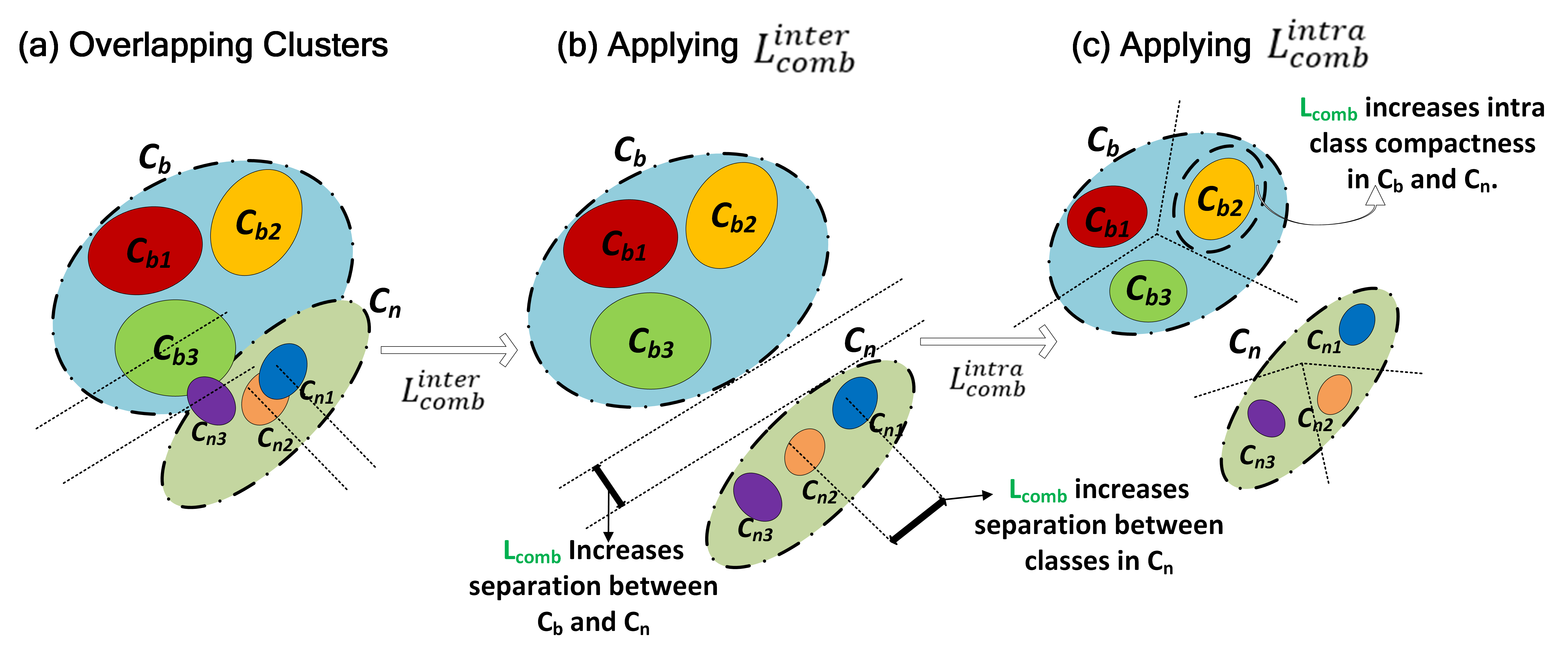}
        \caption{\textbf{Functionality of components in $L_{comb}$} proposed in the \SMILE\ (ours) framework, (a) $L_{comb}^{inter}$ promotes separation between $C_b$ and $C_n$ while (c) $L_{comb}^{intra}$ promotes intra-class compactness.}
        \label{fig:comb_optim_functions}
        \vspace{-2ex}
\end{figure*}

%% file: figures/forgetting_convergence.tex
\begin{figure*}[t]
        \centering
        \includegraphics[width=0.9\textwidth]{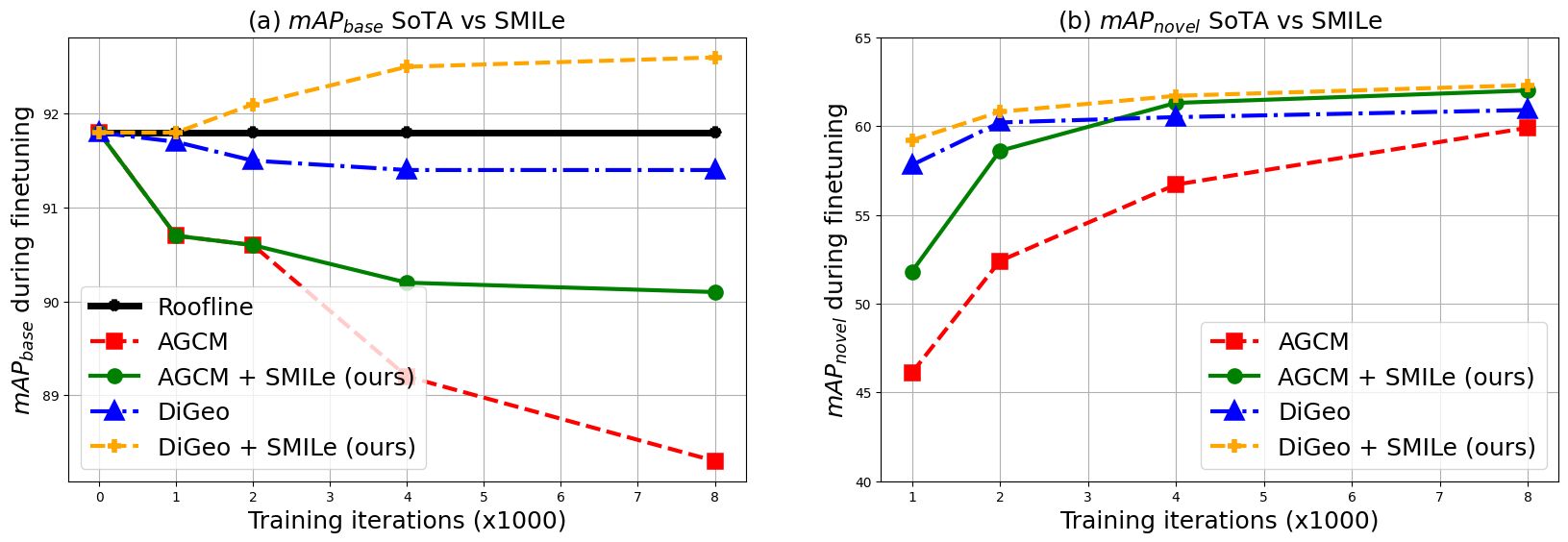}
        \caption{\textbf{Resilience to Catastrophic forgetting and faster convergence in \SMILE} over SoTA approaches. (a) shows that combinatorial losses in \SMILE\ are robust to catastrophic forgetting, while (b) shows that objectives in \SMILE\ results in faster convergence over SoTA FSOD methods (AGCM and DiGeo).}
        \label{fig:forget_converence}
        \vspace{-2ex}
\end{figure*}

%% file: sections/03-related_work.tex
\textbf{Few-Shot Object Detection (FSOD)}: Classical FSOD approaches utilize finetuning \cite{lstd} or distance metric learning \cite{repmet} to adapt features to novel classes. Recent methods employ meta-learning techniques \cite{reweight, addfeat, metarcnn} with episodic training to learn class-specific features. Meta-Reweight \cite{reweight} and Meta-RCNN \cite{metarcnn} use additional feature extractors, while Add-Info \cite{addfeat} leverages feature differences between support and query images. Techniques like \cite{mutualsupport} enhance class-specific features through information sharing, and CME \cite{cme} aims to reduce class confusion. Attention mechanisms \cite{fsod, Zhang_2020_ACCV} are used to identify discriminative features. However, meta-learning approaches are resource-intensive and may fail to generalize to significantly different novel classes. Metric learning strategies like FsDet \cite{fsdet}, FSCE \cite{fscontrastive}, and SRR-FSD \cite{srr} offer better generalization without additional overheads. PNPDet \cite{pnpdet} partially addresses catastrophic forgetting and class confusion. GFSD \cite{Fan_2021_CVPR} proposes a Bias-Balanced RPN to mitigate overfitting in metric learners.

Recent approaches like \cite{labelverifycorrect, identification} adopt weak supervision from unlabelled data or low confidence predictions in RoI pooling layers to generalize to novel classes. These methods often use abundant samples from base classes \cite{digeo} to prevent catastrophic forgetting, adding computational overhead in low-shot settings. Vision transformers \cite{vit} have been adopted in FSOD through methods like imTED \cite{imted} and PDC \cite{pdc}, with reduced computational overhead by using pre-trained attention heads. Alternatively, DiGeo \cite{digeo} and PDC \cite{pdc} learn the geometry or difference in distributions of RoI proposals \cite{fast-rcnn} between object classes to overcome forgetting and confusion. However, these approaches rely on contrastive learning objectives \cite{supcon2020} that struggle to learn discriminative feature embeddings due to adoption of pairwise similarity metrics. Our work, \SMILE, aims to improve the feature learning capacity of existing SoTA approaches, irrespective of their underlying architectures.

\noindent\textbf{Submodular Functions and Combinatorial Objectives} : Submodular functions are recognized as set functions with an inherent diminishing returns characteristic. 
Defined as a set function $f: 2^{\mathcal{V}} \rightarrow \mathbb{R}$ operating on a ground-set $\mathcal{V}$, a function is termed submodular if it adheres to the condition $f(X) + f(Y) \geq f(X \cup Y) + f(X \cap Y), \forall X, Y \subseteq \mathcal{V}$~\cite{fujishige}. 
These functions have garnered considerable attention in research, particularly in fields like data subset selection~\cite{prism}, active learning~\cite{talisman}, and video summarization~\cite{vid_sum_2019, prism} through their ability in modeling concepts such as diversity, relevance, set-cover and representation. 
A subclass of submodular functions, namely Submodular Mutual Information (SMI) functions introduced in \cite{prism} model the similarity and diversity between pairs of object classes establishing itself as a powerful tool to model inter-class bias. 
Recently, Majee et al.~\cite{score} introduces these set-based combinatorial functions as objectives in representation learning and demonstrates their capability in overcoming inter-class bias (by minimizing the similarity between nonidentical object classes) and intra-class variance (maximizing the similarity between instances of the same object class).
However, these functions are yet to be studied in the the context of few-shot learning.
We introduce novel instances of SMI based objectives in \SMILE\ to minimize inter-class bias between base and novel classes. To the best of our knowledge we are the first to introduce novel SMI based combinatorial objectives in conjunction with total information based combinatorial functions through \SMILE\ in a quest to minimize confusion and forgetting in few-shot object detection.

%% file: sections/04-method.tex
\subsection{Problem Definition : Few-Shot Object Detection}
\label{sec:prob_def}
We define a few-shot learner $h(x,\theta)$ as shown in \cref{fig:smile_overview} that receives input data $x$ from base classes $C_{b} \in [1, |C_b|]$ and novel classes $C_{n} \in [1, |C_n|]$ such that $C = \{C_b \cup C_n\}$ and $\{C_b \cap C_n\} = \varnothing$. 
Here, $\theta$ denotes the learnable parameters. The training data can be divided into two distinct parts, base $D_{base}$ and novel $D_{novel}$ such that, $\mathcal{T} = \{D_{base} \cup D_{novel}\}$ and $\{D_{base} \cap D_{novel}\} = \varnothing$. 
\textbf{\SMILE\ introduces a paradigm shift in FSOD by imbibing a combinatorial viewpoint}, where the base dataset, $D_{base} = [A_1^b, A_2^b, \cdots , A_{|C_b|}^b]$, containing abundant training examples from $C_{b}$ base classes and the novel dataset, $D_{novel} = [A_1^n, A_2^n, \cdots, A_{|C_n|}^n]$ containing only $K$-shot ($|A_{i}^n| = K$ for $i \in [1, C_n]$) training examples from $C_n$ novel classes.
\input{figures/overview}
The objective of the few-shot learner \textit{$h(x, \theta)$ is to learn discriminative representation from classes in $D_{novel}$ without degradation in performance on classes in $D_{base}$}. Following FSCE \cite{fscontrastive} we adopt a two-stage training strategy. 
In the \textbf{base training} stage we train $h(x, \theta)$ on abundant samples in $D_{base}$, allowing the model to generalize on the domain of $D_{base}$.
The \textbf{few-shot adaptation} stage adapts $h(x, \theta)$ to \textit{previously unseen} $K$-shot data by fine-tuning on data samples from $D_{base} \cup D_{novel}$ where $|A_k| = K$ for $k \in \{C_b \cup C_n\}$. 
The goal of \SMILE\ is to overcome class confusion and forgetting in FSOD resulting from elevated inter-class bias and intra-class variance as observed in \cite{metadet, majee2021fewshot, agcm}.
The final model $h(x,\theta)$ obtained after two training stages is evaluated on $D_{test}$ containing unseen data samples from both $C_b \cup C_n$.\looseness-1

\subsection{The \SMILE\ Framework}
Adopting a combinatorial viewpoint as disclosed earlier allows us to employ submodular combinatorial functions as learning objectives to tackle confusion and forgetting in FSOD. 
As discussed in \cref{sec:rel_work}, minimizing a Submodular functions naturally models cooperation~\cite{submod_cooperation} while maximizing it models diversity~\cite{submod_diversity} due to their inherent diminishing marginal returns property.
\SMILE\ adopts the aforementioned properties of submodular functions to \textit{define a novel family of combinatorial objective (loss) functions $L_{comb}(\theta)$ which enforces orthogonality in the feature space when applied on Region-of-Interest (RoI) features in FSOD models}.
The loss function $L_{comb}(\theta)$ can be decomposed into two major components - $L_{comb}^{inter}$ minimizes inter-class bias between base and novel classes and $L_{comb}^{intra}$ maximizes intra-class compactness within abundant classes.

For $L_{comb}^{inter}$, \SMILE\ explores a sub-category of combinatorial functions, namely \textit{Submodular Mutual Information} (SMI) which can be defined as $I_f(A_i, A_j) = f(A_i) + f(A_j) - f(A_i \cup A_j)$ \cite{prism, fujishige}, and models the common information between two sets $A_i$ and $A_j$, $\forall i,j \in \mathcal{T}$.
Results in \cite{fujishige, prism} portray $I_f(A_i, A_j; \theta)$ as a measure of the degree of similarity between object classes $A_i$ and $A_j$.
Adopting this definition of SMI, $L_{comb}^{inter}$ \textbf{minimizes the SMI between the base $C_b$ and the novel $C_n$ classes, ensuring sufficient inter-cluster separation} (by minimizing inter-class bias) as shown in \cref{eq:mi_loss}.
$L_{comb}^{inter}$ further \textbf{minimizes the mutual information between classes in $C_n$}, minimizing inter-cluster overlaps between the novel classes. This is visually depicted in \cref{fig:comb_optim_functions}(b) and has been shown to be \textit{effective in mitigating class confusion} in FSOD through our experiments in \cref{sec:experiments}.\looseness-1
\begin{align}
    L_{comb}^{inter}(\theta) = \underset{\substack{b \in C_b \\ n \in C_n}}{\sum} I_f(A_b, A_n; \theta) + \underset{\substack{i, j \in C_n \\ i \neq j}}{\sum} I_f(A_i, A_j; \theta) = \underset{\substack{i \in (C_b \cup C_n) \\ j \in C_n : i \neq j}}{\sum}I_f(A_i, A_j; \theta)
    \label{eq:mi_loss}
\end{align}

In addition to confusion which stems from inter-class bias, \SMILE\ aims at mitigating catastrophic forgetting \cite{majee2021fewshot} in FSOD which has been attributed to large intra-class variance among abundant object classes in \cite{metadet, agcm}.
In coherence to the combinatorial formulation in \SMILE\, we achieve this through $L_{comb}^{intra}$ which minimizes the Total Submodular Information, defined as $S_f(A_1, \cdots, A_{|C|}) = \sum_{k = 1}^{|C|} f(A_k; \theta)$, over sets $A_k \in \mathcal{T}$, given a submodular function $f(A_k; \theta)$.
As discussed earlier, minimizing the submodular information models cooperation which asserts that minimizing $L_{comb}^{inter}$ promotes learning of discriminative feature clusters, penalizing abundant classes to have large feature variance in the embedding space as shown in \cref{fig:comb_optim_functions}(c).
Although submodular functions have been studied in the field of representation learning to minimize intra-class variance in \cite{score}, but primarily differs from \SMILE\ in modeling a longtail recognition task by minimizing the total submodular correlation, which models gain in information when new features are added to a set.
The formulation of $L_{comb}^{intra}$ has been shown in \cref{eq:sim_loss} where we \textbf{minimize the total submodular information within samples in each class in $C_b \cup C_n$} and our experiments in \cref{tab:abl_smi} show the effectiveness of $L_{comb}^{intra}$ in boosting base class performance \textit{asserting the mitigation of catastrophic forgetting}.\looseness-1
\begin{align}
    L_{comb}^{intra}(\theta) = {\underset{b \in C_b}{\sum}} f(A_b, \theta) + {\underset{n \in C_n}{\sum}} f(A_n, \theta) = {\underset{k \in (C_b \cup C_n)}{\sum}} f(A_k, \theta)
    \label{eq:sim_loss}
\end{align}
Ablating on the choice of the submodular function $f$ and SMI functions $I_f$ we introduce several instances of \SMILE\ objectives as discussed in \cref{tab:mi_overview}.\looseness-1

Encapsulating the aforementioned formulations of $L_{comb}^{inter}$ and $L_{comb}^{intra}$ in \SMILE\ we define a joint objective $L_{comb}(\theta)$ which tackles both the challenges of confusion and forgetting. We thus define $L_{comb}(\theta)$ in \cref{eq:info_score_objective} which is the weighted algebraic sum of $L_{comb}^{inter}$ and $L_{comb}^{intra}$ with the weighting factor $\eta$.
\begin{align}
\label{eq:info_score_objective}
\begin{split}
   L_{comb}(\theta) =& (1 - \eta) L_{comb}^{intra}(\theta) + \eta L_{comb}^{inter}(\theta) \\
                    =& {\underset{i \in C_b \cup C_n}{\sum}} \Biggl[(1 - \eta) f(A_i, \theta) + \eta \underset{\substack{j \in C_n \\ i \neq j}}{\sum}I_f(A_i, A_j; \theta) \Biggr]
\end{split}
\end{align}
Note, that the combinatorial objective $L_{comb}(\theta)$ is applied on output features from the RoI Pooling layers in proposal-based~\cite{faster-rcnn, metadet} architectures. To \textbf{promote adoption of \SMILE\ agnostic of the backbone architecture} we introduce a combinatorial head $Z_{comb} = Comb(h, \theta)$ which projects the RoI features to 128-dimensional feature vectors \cite{supcon2020}, $Z_{comb}$ on which $L_{comb}(\theta)$ is applied during the few-shot adaptation stage.

Finally, we summarize the total classification loss in \SMILE\ as depicted in \cref{eq:fsod_cls_loss} as the sum over all three objectives: the classification head $L_{Clf}$, the box regression head $L_{bbox}$ and the combinatorial head $L_{comb}(\theta)$. 
Note that the objectives proposed in \SMILE\ apply only to $Comb(h, \theta)$ while the RoI classification and regression heads are unchanged. This follows the observations in \cite{digeo, metadet} which warrants the boost in performance originating from learning robust feature representations for each RoI predicted by the model.
\begin{align}
   L_{cls}(\theta) =  L_{Clf}(\theta) + L_{bbox}(\theta) + L_{comb}(\theta)
   \label{eq:fsod_cls_loss}
\end{align}

\subsection{Instantiations of $L_{comb}^{inter}$ and $L_{comb}^{intra}$ in the \SMILE\ Framework}
\label{sec:smile_instances}
Given a submodular function $f(A)$ and a Submodular Mutual Information (SMI) function $I_f(A, Q)$ over sets $A$ and $Q$, we derive two instances $L_{comb}^{inter}$ and $L_{comb}^{intra}$ objectives in \SMILE. 
Depending on the choice of $f(A)$ we define two instances: Facility-Location Mutual Information (\SMILE-FLMI) and Graph-Cut Mutual Information (\SMILE-GCMI). 
Inherently, both objectives adopt the cosine similarity metric $S_{ij}(\theta)$ as used in \supcon\ \cite{supcon2020} which can be defined as $S_{ij}(\theta) = \frac{Z_{{comb}_{i}}^{\text{T}} \cdot Z_{{comb}_{j}}}{||Z_{{comb}_{i}}|| \cdot ||Z_{{comb}_{j}}||}$ to compute similarity between sets in the learning objective.
Although the similarity kernel used in \SMILE\ is computed in a pairwise fashion, objectives defined under $L_{comb}$ use it to only compute feature interactions between samples, \textbf{differing from existing approaches in aggregation of pairwise similarities to compute total information and mutual information} over classes in $\mathcal{T}$.\looseness-1

\sloppy
\subsubsection{\SMILE-FLMI} based objective is derived from the Facility-Location Mutual Information (FLMI) \cite{prism} function, expressed as $I_f(Q, A) = \sum_{i \in Q} {\underset{j \in A}{\max}} S_{ij}(\theta) + \lambda\sum_{i \in A} {\underset{j \in Q}{\max}} S_{ij}(\theta)$ and minimizes the maximum similarity (most similar) between sets $Q$ and $A$. 
Given the facility-location (FL) submodular function $f(A, \theta) = {\underset{i \in \mathcal{T}}{\sum}} {\underset{j \in A}{\max}} S_{ij}(\theta)$ over the set $A$, we can derive $L_{comb}^{inter}(\theta)$ and $L_{comb}^{intra}(\theta)$ shown in \cref{eq:fl} as the \textit{\SMILE-FLMI} objective.
Note that $L_{comb}^{inter}(\theta)$ is applied between object classes in $C_b \cup C_n$ and $C_n$ while $L_{comb}^{intra}(\theta)$ is applied over all classes in $C_b \cup C_n$.\looseness-1
\begin{align}
\begin{split}
L_{comb}^{inter}(\theta) &= \underset{\substack{k \in (C_b \cup C_n) \\ l \in C_n : k \neq l}}{\sum} \sum_{i \in A_k}{\underset{j \in A_l}{\max}} S_{ij}(\theta) + \lambda\sum_{i \in A_l} {\underset{j \in A_k}{\max}} S_{ij}(\theta),\\
L_{comb}^{intra}(\theta) &= {\underset{k \in C_b \cup C_n}{\sum}}\sum_{i \in \mathcal{T} \backslash A_k} {\underset{j \in A_k}{\max}} S_{ij}(\theta)   
\end{split}
\label{eq:fl}
\end{align}
Minimizing the $L_{comb}^{inter}$ objective function ensures that the sets $A_l \in C_n$ and $A_k \in C_b \cup C_n$ are disjoint by minimizing the similarity between features in $A_{k}$ and the \textit{hardest negative} ($\sum_{i \in A_k} \max_{j \in A_l} S_{ij}(\theta)$ for $k \in C_b \cup C_n$ and $l \in C_n$) feature vectors in $A_l$. Further, $L_{comb}^{inter}$ enforces sufficient separation between the novel classes themselves to promote learning of disjoint feature clusters even with few-shot data overcoming confusion. Additionally, $L_{comb}^{intra}$ minimizes the total information contained in each set $A_k \in (C_b \cup C_n)$. This objective retains discriminative feature information from each class in $\mathcal{T}$ reducing the impact of forgetting.\looseness-1

\input{tables/loss_func_overview}

\subsubsection{\SMILE-GCMI} based objective described in \cref{eq:gc} minimizes the pairwise similarity of feature vectors between a positive set $A_{k} \in C_b \cup C_n$ and the sets in $A_{l} \in C_n$ while maximizing the similarity between features in each set $A_{k} \in C_b \cup C_n$.
Given two sets $Q$ and $A$, \cite{prism} defines the Graph-Cut SMI to be $I_f(Q, A) = 2 \lambda \sum_{i \in Q} \sum_{j \in A} S_{ij}(\theta)$, where the Graph-Cut function over a set $A$ is given by $f(A, \theta) = \sum_{i \in A}\sum_{j \in \mathcal{T} \setminus A_{k}}S_{ij}(\theta) - \lambda \sum_{i, j \in A} S_{ij}(\theta)$. 
Given the Graph-Cut and the Graph-Cut SMI functions, we derive $L_{comb}^{inter}(\theta)$ and $L_{comb}^{intra}(\theta)$ shown in \cref{eq:gc} as the \textit{\SMILE-GCMI} objective. 
Similar to \SMILE-FLMI, the $L_{comb}^{inter}(\theta)$ is applied between object classes in $C_b \cup C_n$ and $C_n$ while $L_{comb}^{intra}(\theta)$ is applied over all classes in $C_b \cup C_n$.\looseness-1
\begin{align}
\begin{split}
L_{comb}^{inter}(\theta) &= \underset{\substack{k \in (C_b \cup C_n) \\ l \in C_n : k \neq l}}{\sum} 2 \lambda \sum_{i \in A_k} \sum_{j \in A_l} S_{ij}(\theta),\\
L_{comb}^{intra}(\theta) &= {\underset{k \in C_b \cup C_n}{\sum}} \sum_{i \in A_k}\sum_{j \in \mathcal{T} \setminus A_{k}} S_{ij}(\theta) - \lambda \sum_{i, j \in A_k} S_{ij}(\theta)
\end{split}
\label{eq:gc}
\end{align}
Although objectives in \SMILE-FLMI and \SMILE-GCMI are tasked with similar functions, the $L_{comb}^{inter}$ in \SMILE-GCMI minimizes the pairwise similarity between sets in $C_b \cup C_n$ and $C_n$ rather than the most similar set in \SMILE-FLMI. Further, the $L_{comb}^{intra}$ in \SMILE-GCMI scales linearly with size of $A_k$ as described in \cite{score}. This does not allow the model to substantially improve performance on learning discriminative feature representations for classes in both $C_b$ and $C_n$ as the $|A_k| = K$ (number of shots) thus failing to outperform the model trained using \SMILE-FLMI.\looseness-1

The detailed derivations of the aforementioned instances are included in the Supplementary material. 
Our experiments in \cref{sec:ablations} elucidates the fact that \SMILE-FLMI is a better choice to overcome forgetting and confusion in FSOD.

%% file: figures/overview.tex
\begin{figure*}[t]
        \centering
        \includegraphics[width=\textwidth]{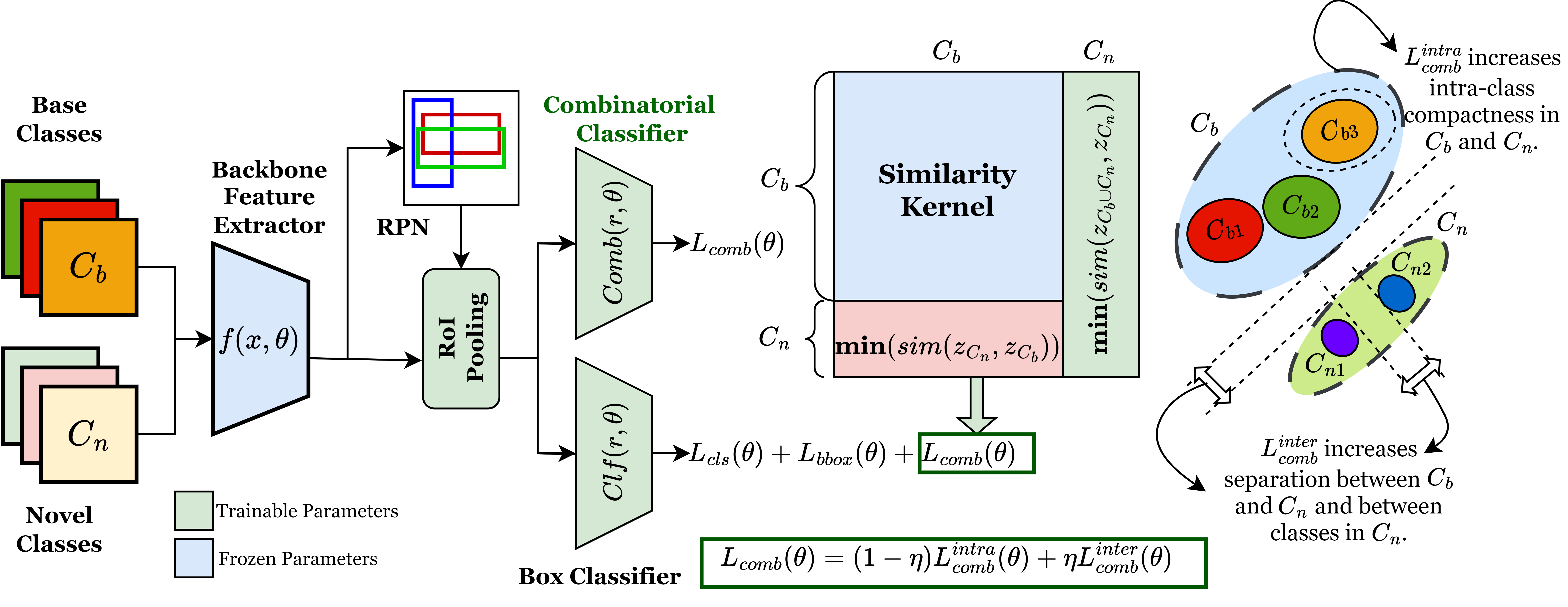}
        \caption{\textbf{Overview of our \SMILE\ framework} highlighting the application of Mutual Information function based objectives in \SMILE\ for the fine-tuning stage of Few-Shot Object Detection.}
        \label{fig:smile_overview}
        \vspace{-2ex}
\end{figure*}

%% file: tables/loss_func_overview.tex
\begin{table*}[t]
      \caption{\textbf{Summary of various instantiations of \SMILE } highlighting the components of the combinatorial objective, $L_{comb}^{inter}$ and $L_{comb}^{intra}$.\looseness-1}
      \centering
      \small
      \resizebox{\textwidth}{!}{\begin{tabular}{l|c|c}
            \hline
            \textbf{Objective}  & 
            \textbf{Instances of $L_{comb}^{inter} (\theta)$}& 
            \textbf{Instances of $L_{comb}^{intra} (\theta)$} \\
            \hline

            

            
            \SMILE-GCMI (ours) &            
            $\underset{\substack{k \in (C_b \cup C_n) \\ l \in C_n : k \neq l}}{\sum} 2 \lambda \sum_{i \in A_k} \sum_{j \in A_l} S_{ij}(\theta)$ &
            ${\underset{k \in C_b \cup C_n}{\sum}} \sum_{i \in A_k}\sum_{j \in \mathcal{T} \setminus A_{k}} S_{ij}(\theta) - \lambda \sum_{i, j \in A_k} S_{ij}(\theta)$ \\

            & & \\ 


            
            \SMILE-FLMI (ours) &
            $\underset{\substack{k \in (C_b \cup C_n) \\ l \in C_n : k \neq l}}{\sum} \sum_{i \in A_k}{\underset{j \in A_l}{\max}} S_{ij}(\theta) + \lambda\sum_{i \in A_l} {\underset{j \in A_k}{\max}} S_{ij}(\theta)$ &
            ${\underset{k \in C_b \cup C_n}{\sum}}\sum_{i \in \mathcal{T} \backslash A_k} {\underset{j \in A_k}{\max}} S_{ij}(\theta)$ \\   
            \hline
      \end{tabular}}\\
      \label{tab:mi_overview}
      \vspace{-2ex}
\end{table*}

%% file: sections/05-experiment.tex
We evaluate models in \SMILE\ by adopting standard evaluation criterion in FSOD \cite{fsdet, reweight} and report the Mean Average Precision ($mAP$) at 50\% Intersection Over Union (IoU) for all our experiments.\looseness-1

\input{tables/pascal_voc}

\subsection{Experimental Setup}
\subsubsection{Datasets}
\label{sec:datasets}
We evaluate our proposed \SMILE\ approach on two few-shot object detection datasets - and PASCAL-VOC \cite{voc} and MS-COCO \cite{mscoco} datasets.\looseness-1

\paragraph{PASCAL-VOC} \cite{voc} dataset consists of 20 classes, out of which 15 are considered as base and 5 as novel classes. The novel classes are chosen at random giving rise to three data splits namely, split-1 (\emph{bird}, \emph{bus}, \emph{cow}, \emph{motorbike}, \emph{sofa}), split-2 (\emph{aeroplane}, \emph{bottle}, \emph{cow}, \emph{horse}, \emph{sofa}) and split-3 (\emph{boat}, \emph{cat}, \emph{motorbike}, \emph{sheep}, \emph{sofa}). Following previous works \cite{reweight}, we use the combined VOC 07+12 datasets for training and evaluate our models on the complete validation set of VOC 2007 for 1, 5, and 10 shot settings.\looseness-1

\paragraph{MS-COCO} \cite{mscoco} dataset consists of 80 classes, out of which 60 are considered as base and 20 as novel classes. Following existing approaches in FSOD \cite{metarcnn} we randomly select 5k samples from $(D_{base} \cup D_{novel})$ to use as the validation set while the remaining samples are used to generate random 10 and 30-shot splits for training of the MS-COCO 2014 dataset. The key difference between VOC and COCO is the large intra-class variance and class-imbalance in COCO.\looseness-1

\subsubsection{Implementation Details}
\label{sec:implementation}
The \SMILE\ framework adopts a architecture agnostic approach and adopt several backbones including Faster-RCNN \cite{faster-rcnn} and ViT \cite{imted}.
For VOC, the input batch size to the network is set to 16 and 2 in the base training and few-shot adaptation stages for Faster-RCNN and ViT based approaches.
The input resolution is set to 764 x 1333 pixels for data splits in COCO, while it is set to 800 x 600 pixels for PASCAL-VOC. 
The hyper-parameters used in the formulation of \SMILE, namely $\eta$ and similarity kernel $S$, are chosen through ablation experiments described in \cref{sec:ablations}.
Results from existing methods are a reproduction of the algorithm from publicly available codebases. 
All our experiments are performed on 4 NVIDIA GTX 1080 Ti GPUs with additional details in the supplementary material and code released at \url{https://github.com/amajee11us/SMILe-FSOD.git}.\looseness-1

\subsection{Results on Few-Shot PASCAL VOC Dataset}
\label{exp:voc}
\Cref{tab:voc} records the results obtained from our \SMILE\ framework on novel splits of the PASCAL-VOC dataset and contrasts it against SoTA FSOD techniques.
We adopt four SoTA approaches FSCE \cite{fscontrastive}, AGCM \cite{agcm}, DiGeo \cite{digeo} and imTED \cite{imted}, covering several backbone architectures Faster-RCNN + FPN (FSCE, AGCM and DiGeo) alongside ViT (imTED, PDC) and introducing \SMILE\ ($M + \SMILE$) approach into existing architectures $M$.
For Faster-RCNN based architectures (FSCE) we show a maximum of 5.7\% (3.3 $mAP$ points) improvement while for FPN based arcitectures (AGCM and DiGeo) we show a 3.5\% (2.1 $mAP$ points for AGCM+\SMILE) improvement.
It is interesting to note that unlike FSCE and AGCM, DiGeo uses abundant samples from $C_b$ alongside few-shot samples in $C_n$ (with upsampling) during finetuning introducing a large inter-class bias. \SMILE\ outperforms DiGeo by up to 2.3 $mAP$ points (split 2, 10-shot) showing the resilience of \SMILE\ towards imbalance, thus overcoming confusion in FSOD.
Additionally, for recently introduced transformer based architectures (imTED + PDC \cite{pdc}) \SMILE\ outperforms the existing SoTA with a maximum improvement of 4.9 $mAP$ points (split 2, 5-shot) thus establishing \SMILE\ as the SoTA on few-shot splits of VOC.
Note, that the choice of objective functions $L_{comb}^{inter}$ and $L_{comb}^{intra}$ for this experiment has been determined to be \SMILE+FLMI through an ablation on the different instances in \SMILE\ as described in \cref{sec:choice_of_smi}.
Finally, \cref{fig:forget_converence}(b) shows that \textbf{introduction of \SMILE\ framework to existing SoTA approaches leads to rapid convergence on the novel classes up to 2x over existing SoTA}. This is significant for mission critical tasks like autonomous driving where the model is required to rapidly learn novel objects to reduce turn-around time.\looseness-1

\subsection{Results on Few-Shot MS-COCO Dataset}
\label{exp:coco}
Similar to the results in PASCAL VOC we demonstrate the results of our \SMILE\ framework on MS-COCO dataset. 
In contrast to VOC, COCO presents an extremely imbalanced setting with a long-tail distribution within $D_{base}$ itself making it really hard for FSOD approaches to achieve SoTA through primitive objective functions.
Following the ablation experiments in \cref{sec:choice_of_smi} we adopt the \SMILE+FLMI objective (best performing) to conduct the experiments on 20 few-shot classes of MS-COCO dataset.
We show that \SMILE\ generalizes existing SoTA approach (imTED + PDC) for COCO dataset by 5.4\% (2.6 $mAP$ points, 30-shot setting). 
This further establishes the generalizability of our approach over varying data distributions (VOC and COCO) while achieving SoTA in FSOD tasks.\looseness-1

\input{tables/ms_coco}

\subsection{Ablation Study}
We conduct ablation experiments on the 10-shot split of VOC (split 1) with hyper-parameters $\eta = 0.5$, cosine similarity metric and $\lambda = 1.0$. Ablations for hyper-parameters are detailed in the supplementary material. 
\label{sec:ablations}
\subsubsection{Components of \SMILE}
\input{tables/ablation_components}
Instantiations in \SMILE\ consists of two main components - $L_{comb}^{inter}$ and $L_{comb}^{intra}$.
We consider three baselines FSCE, AGCM and DiGeo which follow the Faster-RCNN/FPN backbone for this experiment.
First, we introduce $L_{comb}^{intra}$ by adopting the FL based objective as determined through ablation experiments below. 
This objective models intra-class variance and ensures reduction in intra-class variance characterized by boost in base class performance.
Secondly, following the \SMILE-FLMI formulation in \cref{eq:fl} we introduce $L_{comb}^{inter}$ during the few-shot adaptation stage.
Applying this objective minimizes the inter-class bias between $C_b \cup C_n$ and $C_n$, thus improving novel class performance significantly. Nevertheless, we see a slight drop in base class performance due to forgetting prevalent in FSOD tasks.
Finally, we combine both instantiations in \SMILE\ into one single objective as in \cref{eq:info_score_objective} with $\eta=0.5$ and show that \SMILE\ improves both base and novel class performance emerging as the best choice for FSOD.
To demonstrate generalization, we perform this experiment on several SoTA approaches as baseline and show that the results discussed in aforementioned section holds. 
We summarize all the results in \cref{tab:smile_comp}.

\subsubsection{Choice of Combinatorial functions in $L_{comb}$}
\label{sec:choice_of_smi}
\SMILE\ introduces several instances of $L_{comb}^{intra}$ and $L_{comb}^{inter}$. 
To clearly understand the contributions of each of these instances we conduct experiments tabulated in \cref{tab:abl_smi} and determine the best performing formulation which generalizes to existing FSOD architectures.
Unlike other ablation experiments in \cref{sec:ablations}, we conduct our experiments on DiGeo which introduces an extremely imbalanced scenario by using abundant samples in $D_{base}$. We conclude that \SMILE-FLMI which considers the Facility-Location based objective as $L_{comb}^{intra}$ and Facility-Location Mutual Information based objective as $L_{comb}^{inter}$ as the best performing instantiation. This result follows the formulation in \cref{eq:fl} where FL naturally models intra-class compactness in class-imbalanced settings \cite{score} while FLMI penalizes the classes in $C_b$ to learn overlapping feature representations with the classes in $C_n$. \textbf{We use \SMILE-FLMI for all benchmark experiments in \cref{exp:voc} and \cref{exp:coco}}.

\input{tables/ablation_smi}

\subsubsection{Robustness to Catastrophic Forgetting}
\label{sec:forgetting}
One of the most significant challenges in FSOD is the elimination of catastrophic forgetting which manifests as the degradation in the performance of classes in $C_b$ while learning classes in $C_n$. This primarily occurs due to the lack of discriminative feature representations from instances in $D_{base}$ during the few-shot adaptation (stage 2) stage. 
We plot the change in base class performance as the training progresses in existing SoTA methods AGCM and DiGeo against number of training iterations in \cref{fig:forget_converence}(a). 
At first, we contrast the change in base class performance $mAP_{base}$ between AGCM and AGCM+\SMILE\ and observe that AGCM overfits on the few-shot samples in $D_{base}$ reducing the performance on $C_{base}$ as the training progresses. AGCM + \SMILE\ on the other hand better retains the performance on base classes with $\sim$3.5\% better retention in base class performance.
Interestingly, DiGeo is able to retain most of the base class performance with a very small degradation over the roofline (a model trained with only the base classes until convergence). \textbf{Our Digeo+\SMILE\ approach outperforms Digeo by demonstrating base class performance even higher than the roofline} establishing the supremacy of $L_{comb}$ in overcoming inter-class bias and intra-class variance resulting in robustness against catastrophic forgetting.

\subsubsection{Overcoming Class Confusion}
\label{sec:confusion}
\Cref{fig:conf_matrix} highlights the supremacy of the proposed \SMILE\ framework in mitigating class confusion through confusion matrix plots. 
We compare the confusion between classes in $C_b \cup C_n$ of SoTA approaches AGCM and DiGeo before and after introduction of combinatorial objectives in \SMILE.
Although both approaches use K-shot examples for classes in $C_n$, DiGeo differs from AGCM by adopting an upsampling strategy which allows the utilization of abundant examples in $C_b$ while upsampling the instances in $C_n$. 
This injects different degrees of inter-class biases for models trained by adopting AGCM and DiGeo which has been demonstrated as the primary reason for confusion in previous work \cite{majee2021fewshot}.
At first, we observe from \cref{fig:conf_matrix} that by adopting the upsampling based strategy, DiGeo achieves very low confusion between already learnt base classes, leading to significantly lower confusion (5\% among $C_b$ and $C_n$). 
Further, confusion matrix plots in \cref{fig:conf_matrix} show that AGCM+\SMILE\ demonstrates 11\% lower confusion than AGCM and DiGeo+\SMILE\ shows 4\% lower confusion and DiGeo. 
This proves the efficacy of combinatorial objectives ($L_{comb}^{inter}$) in mitigating inter-class bias, thereby reducing confusion between classes. 

\input{figures/conf_matrix}

%% file: tables/pascal_voc.tex
\begin{table*}[t]
      \caption{\small \textbf{Quantitative analysis on PASCAL-VOC dataset:} Few-shot object detection performance ($mAP_{novel}$) on novel class splits of PASCAL-VOC dataset. We tabulate results for K={1, 5, 10} shots from various SoTA techniques in FSOD. * indicates that the results are averaged over 10 random seeds. $\dagger$ indicates a meta-learning strategy (N-way, K-shot training). }
      \centering
      \small
      \scalebox{0.8}{
      \begin{tabular}{l|c|c|ccc|ccc|ccc}
            \hline
            \textbf{Method}          & \multicolumn{1}{p{2cm}|}{\centering Learner \\ Type} & Backbone &  
                                        \multicolumn{3}{c|}{\textbf{Split 1}} &   
                                        \multicolumn{3}{c|}{\textbf{Split 2}} &
                                        \multicolumn{3}{c}{\textbf{Split 3}} \\ 
                                     &&&  K=1 & 5 & 10 &
                                          1   & 5 & 10 &
                                        1   & 5 & 10 \\ 
            \hline \hline
            $\small{\dagger}$ Meta-RCNN \cite{metarcnn}       & Meta      & FRCN-R101 & 19.9  & 45.7  & 51.5  & 10.4  & 34.8  & 45.4  & 14.3  & 41.2  & 48.1  \\
            $\small{\dagger}$Meta-Reweight \cite{reweight}   & Meta      & YOLO V2   & 14.8  & 33.9  & 47.2  & 15.7  & 30.1  & 40.5  & 21.3  & 42.8  & 45.9  \\
            $\small{\dagger}$MetaDet \cite{metadet}          & Meta      & FRCN-R101 & 18.9  & 36.8  & 49.6  & 21.8  & 31.7  & 43.0  & 20.6  & 43.9  & 44.1  \\
            $\small{\dagger}$Add-Info \cite{addfeat}         & Meta      & FRCN-R101 & 24.2  & 49.1  & 57.4  & 21.6  & 37.0  & 45.7  & 21.2  & 43.8  & 49.6  \\
            $\small{\dagger}$CME \cite{cme}            & Meta & YOLO V2 & 17.8   & 44.8  &  47.5 &  12.7 & 33.7   & 40.0   & 15.7  &  44.9 & 48.8  \\
            PNPDet \cite{pnpdet}            & Metric    & DLA-34    & 18.2  & - & 41.0  & 16.6  & -  & 36.4  & 18.9  & -  & 36.2  \\
            FsDet w/ FC \cite{fsdet}        & Metric    & FRCN-R101 & 36.8  & 55.7  & 57.0  & 18.2  & 35.5  & 39.0  & 27.7  & 48.7  & 50.2  \\
            FsDet w/ cos \cite{fsdet}       & Metric    & FRCN-R101 & 39.8  & 55.7  & 56.0  & 23.5  & 35.1  & 39.1  & 30.8  & 49.5  & 49.8  \\
            Retentive-RCNN \cite{rrcnn}    & Metric    & FRCN-R101 & 40.1  & 53.7  & 56.1  & 21.7  & 37.0  & 40.3  & 30.2  & 49.7  & 50.1 \\
            \hline
            FSCE \cite{fscontrastive}       & Metric    & FRCN-R101 & 41.0 &  57.4 & 57.8 & 27.3	& 44.4 & 49.8 & 40.1	& 53.2	& 57.7  \\
            \rowcolor{LightGreen}
            \textbf{FSCE + \SMILE\ (ours)}     & \textbf{Comb.} & \textbf{FRCN-R101} & \textbf{41.2} & \textbf{57.9} & \textbf{61.1} & \textbf{29.2} & \textbf{44.6} & \textbf{50.5} & \textbf{41.3} & \textbf{55.6}& \textbf{59.0}  \\
            \hline
            AGCM \cite{agcm}                & Metric    & FRCN-R101 & 40.3 & 58.5  & 59.9 &	27.5    & 49.3 & 50.6 &	42.1    & 54.2  & 58.2 \\
            \rowcolor{LightGreen}
            \textbf{AGCM + \SMILE\ (ours)}     & \textbf{Comb.} & \textbf{FRCN-R101} & \textbf{40.9} & \textbf{59.7}  & \textbf{62.0} & \textbf{31.9} & \textbf{49.5} & \textbf{52.3} & \textbf{42.6} & \textbf{56.4} & \textbf{61.4} \\
            \hline
            DiGeo \cite{digeo}              & Metric    & FRCN-R101               & 36.0 & 54.1  & 60.9 & 20.7 & 42.8 & 47.1 & 27.5 & 47.3 & 52.9 \\
            \rowcolor{LightGreen}
            \textbf{DiGeo + \SMILE (ours)}    & \textbf{Comb.} & \textbf{FRCN-R101} & \textbf{36.1} & \textbf{56.6}  & \textbf{62.3} & \textbf{26.5} & \textbf{44.1} & \textbf{47.3} & \textbf{33.1} & \textbf{51.9} & \textbf{56.4}  \\
            \hline 
            imTED \cite{imted}              & Metric        & ViT-B          & 31.9 & 71.9 & 77.0 &	22.7 & 52.2 & 57.7 & 12.6 & 69.6 & 72.8 \\
            imTED + PDC \cite{pdc}          & Metric        & ViT-B          & \textbf{36.6} & 73.1 & 77.1 & 15.5 & 51.8 & 56.0 & \textbf{18.9} & 67.9 & 72.8 \\
            \rowcolor{LightGreen}
            \textbf{PDC + \SMILE\ (ours)}      & \textbf{Comb.}& \textbf{ViT-B} & \textbf{36.6} & \textbf{75.2} & \textbf{77.9} & \textbf{27.1} & \textbf{52.7} & \textbf{58.3} &	15.1 & \textbf{70.0} & \textbf{74.7} \\
            \hline
      \end{tabular}
      \label{tab:voc}}
\end{table*}

%% file: tables/ms_coco.tex
\begin{table}[t]
\label{tab:coco}
\centering
\small
\caption{\textbf{Performance of \SMILE\ on MS COCO dataset} : Our \SMILE\ objectives demonstrate better generalizability while outperforming SoTA FSOD approaches on novel class performance $mAP_{50}$ (novel).}
\scalebox{0.9}{\begin{tabular}{c|ccc|ccc}
    \hline
     \multirow{2}{*}{\textbf{Method}} & \textbf{$mAP$} & \textbf{$mAP_{50}$} & \textbf{$mAP_{75}$} & \textbf{$mAP$} & \textbf{$mAP_{50}$} & \textbf{$mAP_{75}$} \\
    \cline{2-7}
    & \multicolumn{3}{c|}{\textbf{10-shot}}  & \multicolumn{3}{c}{\textbf{30-shot}} \\ 
    \hline \hline
    Meta-Reweight \cite{reweight} & 5.6 & 12.3 & 4.6   & 9.1 & 19.0 & 7.6\\
    Meta-RCNN \cite{metarcnn}      & 8.7 & 19.1 & 6.6   & 12.4 & 25.3 & 10.8\\
    TFA w/cos \cite{fsdet}        & 10.0 & - & 9.3     & 13.7 & - & 13.4 \\
    Add-Info \cite{addfeat}        & 12.5 & 27.3 & 9.8  & 14.7 & 30.6 & 12.2 \\
    MPSR \cite{wu2020mpsr}         & 9.8 & 17.9 & 9.7   & 14.1 & 25.4 & 14.2\\
    FSCE \cite{fscontrastive}     & 11.9 & - & 10.5    & 16.4 & - & 16.2 \\
    FADI \cite{fadi}              & 12.2 & 22.7 & 11.9 & 16.1 & 29.1 & 15.8 \\
    CME \cite{cme}                & 15.1 & 24.6 & 16.4 & 16.2 & - & - \\
    FCT \cite{fct}                & 17.1 & 30.2 & 17.0 & 21.4 & 35.5 & 22.1\\
    \hline
    imTED-B \cite{imted}          & 22.5 & 36.6 & 23.7 & 30.2 & 47.4 & 32.5 \\
    imTED-B+PDC \cite{pdc}        & 23.4 & 38.1 & 24.5 & 30.8 & 47.3 & 33.5 \\
    \rowcolor{LightGreen}
    \textbf{PDC + \SMILE\ (ours)}  & \textbf{25.8} & \textbf{40.1} & \textbf{26.1} & \textbf{31.0} & \textbf{49.9} & \textbf{33.6}\\
    \hline
\end{tabular}}
\end{table}

%% file: tables/ablation_components.tex
\begin{table}[t]
      \caption{\small Ablation on various components of the proposed \SMILE\ approach. }
      \small
      \centering
      \scalebox{0.9}{
        \begin{tabular}{l|ccc|cc}
            \hline
            \multirow{2}{*}{Method} & \multirow{2}{*}{Baseline} & $f(A_i)$ & $I_{f}(A_i, A_j)$ & \multirow{2}{*}{$mAP_{base}$} & \multirow{2}{*}{$mAP_{novel}$} \\ 
                                    &  & ($L_{comb}^{intra}$) & ($L_{comb}^{inter}$) &  &  \\
            \hline \hline
            FsDet w/ cos            & -          &  -         &  -          &   23.6  & 39.8     \\
            \hline
            \multirow{4}{*}{FSCE}   & \checkmark &            &             &   86.1  & 57.8     \\
                                    & \checkmark & \checkmark &             &   89.6  & 60.1     \\
                                    & \checkmark &            & \checkmark  &   88.3  & 61.0     \\
                                    & \checkmark & \checkmark & \checkmark  &   \textbf{89.8}  & \textbf{61.1}     \\
            \hline
            \multirow{4}{*}{AGCM }  & \checkmark &            &             &   87.6  & 58.0     \\
                                    & \checkmark & \checkmark &             &   88.6  & 61.3  \\
                                    & \checkmark &            & \checkmark  &   88.9  & 61.8  \\
                                    & \checkmark & \checkmark & \checkmark  &   \textbf{89.3}  & \textbf{61.8}  \\
            \hline
            \multirow{4}{*}{DiGeo}  & \checkmark &            &             &   90.5  & 60.9  \\
                                    & \checkmark & \checkmark &             &   92.3  & 61.7  \\
                                    & \checkmark &            & \checkmark  &   91.4  & 62.0  \\
                                    & \checkmark & \checkmark & \checkmark  &   \textbf{92.6}  & \textbf{62.3}  \\
            \hline
      \end{tabular}
      }
      \label{tab:smile_comp}
\end{table}

%% file: tables/ablation_smi.tex
\begin{table}[t]
\caption{\small Ablation on the choice of Submodular Information function $I_f$ and Submodular Information function $f$ for $L_{comb}$ in \SMILE.}
\label{tab:abl_smi}
\centering
\small
\scalebox{0.8}{\begin{tabular}{l|cc|cc}
\hline
Model & $f(A, \theta)$ & $I_f(A_i, A_j)$ & $mAP_{base}$ & $mAP_{novel}$ \\ \hline \hline
\multirow{3}{2cm}{\centering DiGeo \cite{digeo} }     
                              & -           & -              & 87.9          & 57.4 \\
                              & GC          & GCMI           & 92.6          & 60.9 \\
                              & \textbf{FL} & \textbf{FLMI} & \textbf{93.1} & \textbf{62.3} \\ \hline  
\multirow{3}{2cm}{\centering FSCE \cite{fscontrastive} }     
                              & -           & -              & 86.1          & 57.8 \\
                              & GC          & GCMI           & 87.4          & 60.3 \\
                              & \textbf{FL} & \textbf{FLMI} & \textbf{89.8} & \textbf{61.1} \\ \hline  
\end{tabular}}
\end{table}

%% file: figures/conf_matrix.tex
\begin{figure*}[t]
        \centering
        \includegraphics[width=\textwidth]{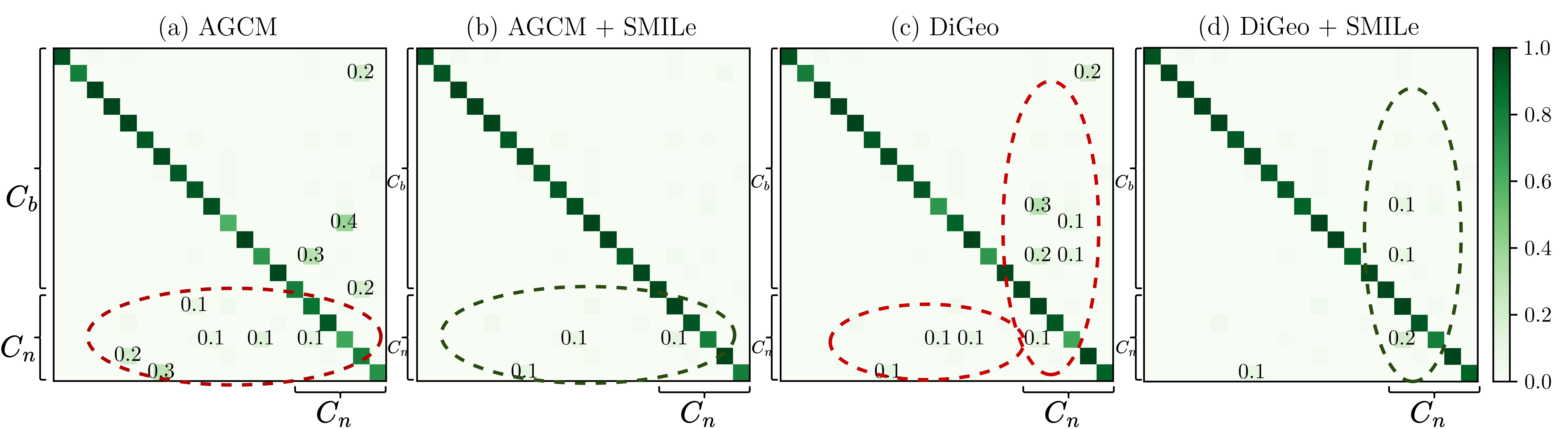}
        \caption{\textbf{Ablation on Overcoming Class Confusion in \SMILE}. (a,b) \SMILE\ demonstrates 11\% lower confusion over AGCM and (c,d) 4\% lower confusion over DiGeo. Only significant numbers are highlighted. Best viewed in 200\% zoom. \looseness-1}
        \label{fig:conf_matrix}
        \vspace{-4ex}
\end{figure*}

%% file: sections/06-conclusion.tex
In this work, we have presented a novel approach to Few-Shot Object Detection (FSOD) by introducing a combinatorial viewpoint through the \textbf{\SMILE} framework. 
By leveraging the properties of set-based combinatorial functions, \SMILE\ aims to address the challenges of class confusion and catastrophic forgetting, which are prevalent in FSOD tasks. Our approach incorporates Submodular Mutual Information (SMI) and Submodular Information Measures (SIM) to penalize overlapping features between base and novel classes and to ensure the formation of compact feature clusters, respectively. 
The experimental results on PASCAL-VOC and MS-COCO benchmarks demonstrate the effectiveness of \SMILE, showing significant improvements in novel class performance, faster convergence, and a reduction in class confusion and catastrophic forgetting. Overall, \SMILE\ offers a promising direction for advancing the state-of-the-art in FSOD by providing a generalized framework that is adaptable to various underlying architectures and capable of handling the complexities associated with few-shot learning in object detection.

%% file: sections/07-supplementary.tex
\section{Notation}
\label{app:notation}
Following the problem definition in Sec. 3.1 we introduce several important notations in Table \ref{tab:notations} that are used throughout the paper. 
\input{tables/notations}

\section{Implementation Details}
As discussed in the main paper the \SMILE\ framework proposes an architecture agnostic approach and adopts several backbones - Faster-RCNN \cite{faster-rcnn} and ViT \cite{imted, pdc}.
We conduct experiments on PASCAL-VOC \cite{pascalvoc} and COCO \cite{mscoco} datasets.
For VOC, the input batch size to the network (both Faster-RCNN and ViT based approaches) is set to 16 and 2 in the base training and few-shot adaptation stages for Faster-RCNN and ViT based approaches.
Our experiments in Tab. 2 of the main paper applies the combinatorial formulation in \SMILE\ to four different architectures - FSCE \cite{fscontrastive}, AGCM \cite{agcm}, DiGeo \cite{digeo} and imTED+PDC \cite{pdc}.

For FSCE and AGCM we train the model for a maximum of 12k iterations and 6k iterations respectively with an initial learning rate of 0.01 with a batch size of 16 for VOC and 8 for COCO datasets. 
For DiGeo and DiGeo+\SMILE\ we train the model for 15k steps with 200 warmup steps with a batch size of 8 and an initial learning rate of 0.05 for both datasets. The codebase for AGCM + \SMILE\ and FSCE + \SMILE\ has been released at \url{https://github.com/amajee11us/SMILe-FSOD.git}. 
For the DiGeo + \SMILE\ architecture we follow the authors in \cite{digeo} and introduce $Comb(h, \theta)$ in the \textit{distill} stage of the training process. 
Following the authors in \cite{digeo} we use abundant samples of the base classes and K-shot (few-shot) samples of the novel classes and use the same set of hyper-parameters as released in our codebase at \url{https://github.com/amajee11us/SMILe-FSOD/tree/digeo}.

Due to adoption of ViT \cite{vit} based architecture in imTED + PDC and imTED + PDC + \SMILE\ architectures, we train the model with a batch size of 2 (as used in \cite{pdc}) with an initial learning rate of 1e-4 for a total of 108 epochs with a step learning rate scheduler. We release the code for training and inferencing on the PDC + \SMILE\ is released at \url{https://github.com/amajee11us/SMILe-FSOD/tree/pdc_SMILe}. 

The $Comb(h, \theta)$ architecture is applied only during the few-shot adaption stage (across architectures) of model training and the input resolution is set to 764 x 1333 pixels for data splits in COCO, while it is set to 800 x 600 pixels for PASCAL-VOC.
For all architecture variants we adopt the Stronger Baseline introduced in FSCE \cite{fscontrastive} with a trainable Region Proposal Network (RPN) and RoI Pooling layer alongside increasing the number of RoI proposals to 2048 (double the number as compared to \cite{fsdet}). The additional RoI proposal features help capture the low confidence novel classes in the initial training iterations leading to faster convergence.
Additionally we introduce two hyper-parameters in the formulation of \SMILE, namely $\eta$ and similarity kernel $S$, are chosen through ablation experiments described in \cref{sec:abl_hyperparameters}.
Following existing research \cite{fscontrastive, agcm} we report the novel class performance for 1, 5, 10 shot settings for VOC and 10, 30 settings for COCO averaged over 10 distinct seeds\footnote{The default seeds for all our experiments were adapted from \url{http://dl.yf.io/fs-det/datasets/}}.
Results from existing methods are a reproduction of the algorithm from publicly available codebases.\looseness-1

\section{Ablation : Hyper-Parameters in \SMILE}
\label{sec:abl_hyperparameters}
\input{tables/ablation_hyperparameter}
We perform ablation on various hyper-parameters introduced in \SMILE\ and derive their values which lead to the best possible base and novel class performance in the few-shot adaptation stage. 
For all our experiments we consider the AGCM \cite{agcm} architecture as the baseline and train and evaluate the model on the PASCAL VOC dataset.
\SMILE\ introduces two important hyper-parameters, similarity kernel $S$ and $\eta$. 
The choice of similarity kernel determines how gradients are calculated in the objective function and they magnitude of $S$ depends on the model parameters $\theta$. We chose the cosine similarity (indicated as Cosine in \cref{tab:hyper}) metric over others as it achieves the best overall performance.
The hyper-parameter $\eta$ controls the contribution of $L_{comb}^{inter}$ over $L_{comb}^{intra}$ such that their overall contributions add up to 1.0 (100\%). We vary the value for $\eta$ between $\alpha = 0.0$ to $\alpha =  1.0$ and record the variation in performance of the novel classes in \cref{tab:hyper}. We choose $\eta = 0.5$ for our experiments across all datasets.

Additionally, we introduce the hyper-parameter $\lambda$ specific to \SMILE-GCMI to control the degree of compactness of the feature cluster ensuring sufficient diversity is maintained in the feature space. Experimental results in \cref{tab:hyper} indicates that $\lambda \geq 1.0$ is necessary for Graph-Cut in $L_{comb}^{intra}$ to be submodular thus we adopt $\lambda=1.0$ for our experiments.

\section{Proofs for Theorems in \SMILE}
In this section, we provide the necessary proofs leading to the derivation of the components of $L_comb$ namely, $L_{comb}^{inter}$ and $L_{comb}^{intra}$ for different instantiations of the submodular function $f(A, \theta)$ over any given set $A$. We restate the theorems as in the main paper for better readability.

\subsection{Derivation of \SMILE-FLMI}
\sloppy
Given $I_f(Q, A) = \sum_{i \in Q} {\underset{j \in A}{\max}} S_{ij}(\theta) + \lambda\sum_{i \in A} {\underset{j \in Q}{\max}} S_{ij}(\theta)$ and $f(A, \theta) = {\underset{i \in \mathcal{T}}{\sum}} {\underset{j \in A}{\max}} S_{ij}(\theta)$ representing the facility-location mutual information function and facility-location submodular function respectively over sets $A$ and $Q$ then, we derive the expressions for \SMILE-FLMI as a summation of $L_{comb}^{inter}(\theta)$ and $L_{comb}^{intra}(\theta)$ respectively as depicted in Eq. 6 of the main paper.

Lets first derive the $L_{comb}^{intra}$ from the total information formulation given $f(A, \theta)$ as the underlying submodular function.  
From the definition of $L_{comb}^{intra}$, the objective can be derived as $L_{comb}^{intra}(\theta) = {\underset{k \in (C_b \cup C_n)}{\sum}} f(A_k, \theta)$. Substituting the instance of FL $f(A, \theta) = {\underset{i \in \mathcal{V}}{\sum}} {\underset{j \in A}{\max}} S_{ij}(\theta)$ in the equation we get:
\begin{align*}
        L_{comb}^{intra}(\theta) &= \overset{|C_b \cup C_n|}{\underset{k=1}{\sum}} f(A_k, \theta) \\
                        &= {\underset{k \in (C_b \cup C_n)}{\sum}} {\underset{i \in \mathcal{T}}{\sum}} {\underset{j \in A_k}{\max}} S_{ij}(\theta) \\
                        &= {\underset{k \in (C_b \cup C_n)}{\sum}} {\underset{i \in \mathcal{T} \setminus A_k}{\sum}} {\underset{j \in A_k}{\max}} S_{ij}(\theta) + {\underset{k \in (C_b \cup C_n)}{\sum}} {\underset{i \in A_k}{\sum}} {\underset{j \in A_k}{\max}} S_{ij}(\theta) \\
        L_{comb}^{intra}(\theta) &= {\underset{k \in (C_b \cup C_n)}{\sum}} {\underset{i \in \mathcal{T} \setminus A_k}{\sum}} {\underset{j \in A_k}{\max}} S_{ij}(\theta) + |\mathcal{T}|\\                        
\end{align*}
\sloppy
Here, ${\underset{i \in A_k}{\sum}} {\underset{j \in A_k}{\max}} S_{ij}(\theta)$ is a constant over the set $A_k$. 
Hereafter, we provide the proof for the $L_{comb}^{inter}$ formulation which can be derived from $L_{comb}^{inter}(\theta) = \underset{\substack{i \in (C_b \cup C_n) \\ j \in C_n : i \neq j}}{\sum} I_f(A_i, A_j; \theta)$. Given the Submodular Mutual Information function $I_f(Q, A) = \sum_{i \in Q} {\underset{j \in A}{\max}} S_{ij}(\theta) + \lambda\sum_{i \in A} {\underset{j \in Q}{\max}} S_{ij}(\theta)$ over two distinct sets $Q$ and $A$, we substitute the value of $I_f$ in $L_{comb}^{inter}$.
\begin{align*}
        L_{comb}^{inter}(\theta) &= \underset{\substack{k \in (C_b \cup C_n) \\ l \in C_n : k \neq l}}{\sum} I_f(A_k, A_l; \theta)  \\
                        &= {\underset{k \in (C_b \cup C_n)}{\sum}} {\underset{\substack{l \in C_n \\ k \neq l}}{\sum}}  \Biggl[{\underset{i \in A_k}{\sum}}{\underset{j \in A_l}{\max}} S_{ij}(\theta) + \lambda {\underset{i \in A_l}{\sum}}{\underset{j \in A_k}{\max}} S_{ij}(\theta) \Biggr]\\
        L_{comb}^{inter}(\theta) &= \underset{\substack{k \in (C_b \cup C_n) \\ l \in C_n : k \neq l}}{\sum}  \Biggl[{\underset{i \in A_k}{\sum}}{\underset{j \in A_l}{\max}} S_{ij}(\theta) + \lambda {\underset{i \in A_l}{\sum}}{\underset{j \in A_k}{\max}} S_{ij}(\theta) \Biggr]\\                       
\end{align*}
Note that the similarity computed between sets of features depend on the parameters of the model $\theta$. 
We parallelized the computation of \SMILE-FLMI in our implementation using vectorized calculations available in the Pytorch (\url{https://pytorch.org/}) library.

\subsection{Derivation of \SMILE-GCMI}
From $I_f(Q, A) = 2 \lambda \sum_{i \in Q} \sum_{j \in A} S_{ij}(\theta)$ and $f(A, \theta) = \sum_{i \in A}\sum_{j \in \mathcal{T} \setminus A}S_{ij}(\theta) - \lambda \sum_{i, j \in A} S_{ij}(\theta)$ representing the Graph-Cut (GC) mutual information function and Graph-Cut submodular function respectively over sets $A$ and $Q$ then, $L_{comb}^{inter}(\theta)$ and $L_{comb}^{intra}(\theta)$  we derive the expressions for \SMILE-GCMI as a summation of $L_{comb}^{inter}(\theta)$ and $L_{comb}^{intra}(\theta)$ respectively as depicted in Eq. 7 in the main paper.

From the definition of $f(A, \theta)$, the \SMILE-GCMI ($L_{comb}^{intra}$) objective can be derived by substituting the instance of GC $f(A_k, \theta)$ in the equation we get:    
\begin{align*}
        L_{comb}^{intra}(\theta) &= {\underset{k \in (C_b \cup C_n)}{\sum}} f(A_k, \theta) \\
                        &= {\underset{k \in (C_b \cup C_n)}{\sum}} \sum_{i \in A_k}\sum_{j \in \mathcal{T}}S_{ij}(\theta) - \lambda \sum_{i, j \in A_k} S_{ij}(\theta) \\ 
                        &= {\underset{k \in (C_b \cup C_n)}{\sum}} {\underset{\substack{i \in A_k \\ j\in \mathcal{T} \setminus A_k}}{\sum}} S_{ij}(\theta) + {\underset{k \in (C_b \cup C_n)}{\sum}} {\underset{\substack{i \in A_k \\ j\in A_k}}{\sum}} S_{ij}(\theta) - \lambda \sum_{i, j \in A_k} S_{ij}(\theta) \\
\end{align*}
Here, the term ${\underset{k \in (C_b \cup C_n)}{\sum}} {\underset{i \in A_k, j\in A_k}{\sum}} S_{ij}(\theta)$ represents a sum of pairwise similarities over all sets in $\mathcal{V}$. Thus, its value is a constant for a fixed training/ evaluation dataset. Using this condition and ignoring the constant term, we can show that:
\begin{align*}
        L_{comb}^{intra}(\theta) &= {\underset{k \in (C_b \cup C_n)}{\sum}} {\underset{i \in A_k, j\in \mathcal{T} \setminus A_k}{\sum}} S_{ij}(\theta) - \lambda \sum_{i, j \in A_k} S_{ij}(\theta) \\
\end{align*}

Hereafter, we provide the proof for the $L_{comb}^{inter}$ formulation which can be derived from $L_{comb}^{inter}(\theta) = \underset{\substack{i \in (C_b \cup C_n) \\ j \in C_n : i \neq j}}{\sum} I_f(A_i, A_j; \theta)$. Given the Submodular Mutual Information function $I_f(Q, A) = 2 \lambda \sum_{i \in Q} \sum_{j \in A} S_{ij}(\theta)$ over two distinct sets $Q$ and $A$, we substitute the value of $I_f$ in $L_{comb}^{inter}$.
\begin{align*}
        L_{comb}^{inter}(\theta) &= \underset{\substack{k \in (C_b \cup C_n) \\ l \in C_n : k \neq l}}{\sum} I_f(A_k, A_l; \theta)  \\
                        &= {\underset{k \in (C_b \cup C_n)}{\sum}} {\underset{\substack{l \in C_n \\ k \neq l}}{\sum}} \Biggl[2 \lambda \sum_{i \in A_k} \sum_{j \in A_l} S_{ij}(\theta) \Biggr]\\
        L_{comb}^{inter}(\theta) &= \underset{\substack{k \in (C_b \cup C_n) \\ l \in C_n : k \neq l}}{\sum}  2 \lambda  \sum_{\substack{i \in A_k \\ j \in A_l}} S_{ij}(\theta)\\                       
\end{align*}
From the above formulation, we observe that \SMILE-GCMI is computationally inexpensive as compared to \SMILE-FLMI, but our experimental results show that \SMILE-FLMI outperforms \SMILE-GCMI. 
This is predominantly because the objective function in \SMILE-FLMI scales non-linearly with the size of the set $|A_k|$ inherently modelling the imbalance between the already learnt classes $C_b$ and the newly added ones $C_n$.

\input{figures/qualitative_voc}

\section{Ablation : Qualitative Results from \SMILE\ Against SoTA}
\label{sec:abl_qualitative}
Figure \ref{fig:qual} shows qualitative results for our proposed \SMILE\ method on the PASCAL VOC \cite{pascalvoc}.
Due to limited compute resources we conduct experiments on FSCE and AGCM approaches before and after introduction of the \SMILE\ approach.
\Cref{fig:qual}(a) shows that introduction of \SMILE\ is resilient to scale (varying sizes) and occlusion, while \Cref{fig:qual}(c) shows significant base class forgetting in both FSCE and AGCM.
\Cref{fig:qual}(b) shows significant catastrophic forgetting in FSCE and AGCM which has also been shown to be overcome by \SMILE\ while \cref{fig:qual}(d) demonstrate resilience against color and texture variations.
Overall, \SMILE\ handles forgetting and confusion significantly over SoTA approaches while minimizing the degradation in performance of the base classes.



\section{Limitations and Future Work}
From the experiments proposed in our paper, we demonstrate the generalizability as well as the supremacy of our approach in handling class confusion and forgetting.
Although, significant progress has been demonstrated in overcoming confusion and forgetting by \SMILE\, some amount of confusion and forgetting continue to plague this domain. 
This would definitely be a direction for future research both in FSOD and in combinatorial representation learning.
Further, \SMILE\ demonstrates success in the 5/10 shot setting, we observe suboptimal performance in the 1-shot case. 
This is a plausible direction that the authors would be studying in depth in the near future.
In the current setting, novel classes need to be first labelled by human annotators before being served to the \SMILE\ framework. Unfortunately, to rapidly adapt to the open-world setting our model should be able to generalize to unknown Region-of-Interests, which the authors would like to study in future research.

%% file: tables/notations.tex
\begin{table*}[ht]
      \caption{Collection of notations used in the paper.}
      \centering
       \scalebox{0.9}{\begin{tabular}{ c | c }
            \hline
           \textbf{Symbol}  & \textbf{Description} \\
            \hline
            $\mathcal{T}$ & The training Set. $|\mathcal{T}|$ denotes the size of the training set. \\
            $h(x, \theta)$ & Feature extractor without the box classifier and regressor. \\
            $Clf(.,\theta)$ & Multi-Layer Perceptron as classifier and regression head (as in Faster-RCNN). \\
            $Comb(.,\theta)$ & Multi-Layer Perceptron as Combinatorial Classifier head. \\
            $\theta$ & Parameters of the feature extractor. \\
            $S_{ij}(\theta)$ & Similarity between images $i,j \in \mathcal{T}$. \\
            $C_b$ & Classes indexed in the base dataset. \\
            $C_n$ & Classes indexed in the novel dataset. \\
            $C$ & All classes in the input dataset $\mathcal{T}$ represented as $C_b \cup C_n$.\\
            $A_k$ & Target set containing feature representation from a single class $k \in C$.\\
            $f(A)$ & Submodular Information function over a set $A$. \\
            $I_{f} (A, Q)$ & Mutual information function between set $A$ and $Q$. \\
            $L(\theta)$ & Loss value computed over all classes $i \in C$. \\
            $L_{comb}$ & Combiatorial Objectives in \SMILE. \\
            \hline
      \end{tabular}}\\
      \label{tab:notations}
      
\end{table*}

%% file: tables/ablation_hyperparameter.tex
\begin{table}[t]
\caption{\small Ablation study for the key hyper-parameters in \SMILE. The chosen values are \underline{underlined} and associated performance values are indicated in \textbf{bold}.}
\label{tab:hyper}
\centering
\small
\scalebox{0.8}{
\begin{tabular}{c|c|c|c}
\hline
Parameter & Value & $mAP_{base}$ & $mAP_{novel}$ \\ \hline \hline
\multirow{3}{2cm}{\centering Similarity \\ Kernel (S) }     
                              & Euclidean                 & 84.7                         & 59.4                          \\
                              & \underline{Cosine}                    & \textbf{88.9}                & \textbf{61.3}                          \\
                              & RBF                       & 86.1                         & 59.6                          \\ \hline
\multirow{6}{2cm}{\centering $\eta$ \\(sim. Kernel = Cosine)}
                              & 0.0                       & 87.5                         & 59.9                          \\
                              & 0.2                       & 88.7                         & 60.2                          \\
                              & \underline{0.5}                       & \textbf{89.3}                & \textbf{62.0}                          \\
                              & 0.8                       & 86.7                         & 61.1                          \\
                              & 1.0                       & 86.1                         & 58.3  \\ \hline                   
\multirow{5}{2cm}{\centering $\lambda$ \\ $L_{comb}^{inter}$ (S = Cosine, $\eta$=0.5)}
                              & 0.5                       & 82.1                         & 57.3                          \\
                              & 0.7                       & 86.4                         & 60.1                          \\
                              & \underline{1.0}                       & \textbf{87.4}                & \textbf{60.3}                          \\
                              & 1.2                       & 87.4                         & 59.9                          \\
                              & 1.5                       & 87.1                         & 54.6                          \\ \hline     
\end{tabular}}
\end{table}

%% file: figures/qualitative_voc.tex
\newcommand{\centered}[1]{\begin{tabular}{l} #1 \end{tabular}}

\begin{figure*}
        \centering
        \begin{tabular}{lcccc}
            \rotatebox{90}{\small AGCM \cite{agcm}} &
                \includegraphics[width=0.23\textwidth]{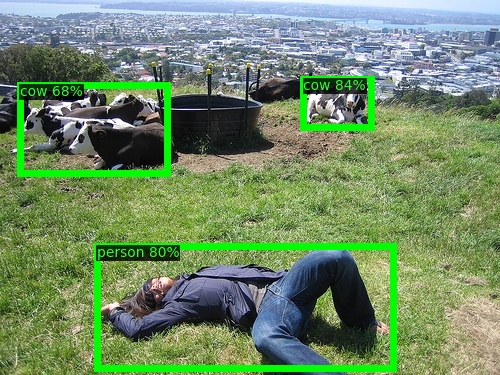} & 
                \includegraphics[width=0.23\textwidth]{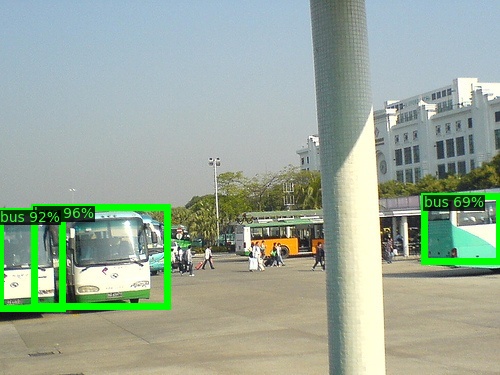} & 
                \includegraphics[width=0.23\textwidth]{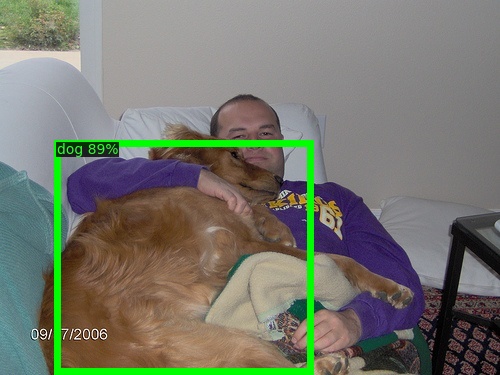} &
                \includegraphics[width=0.23\textwidth]{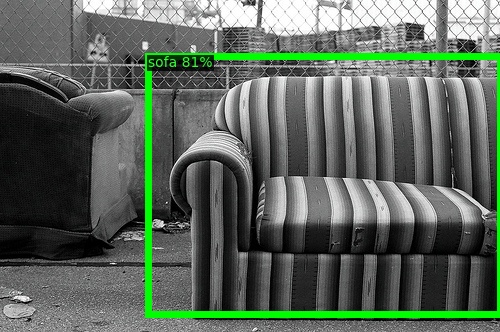} \\
            \rotatebox{90}{\small AGCM + \SMILE\ } & 
                \includegraphics[width=0.23\textwidth]{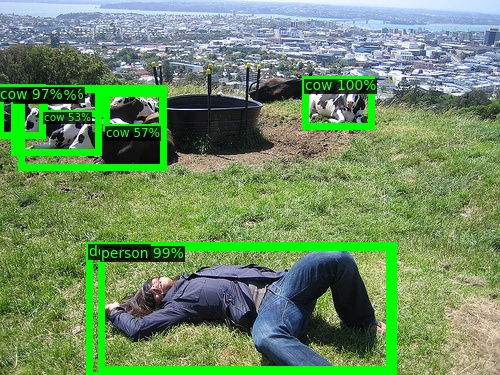} & 
                \includegraphics[width=0.23\textwidth]{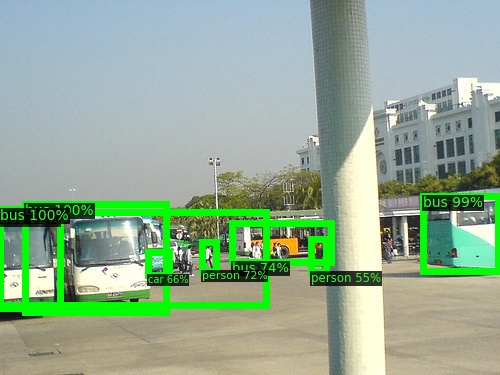} & 
                \includegraphics[width=0.23\textwidth]{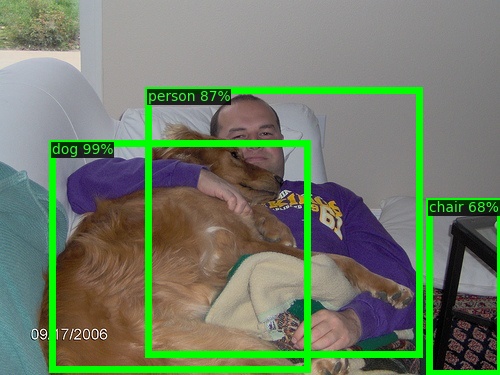} &
                \includegraphics[width=0.23\textwidth]{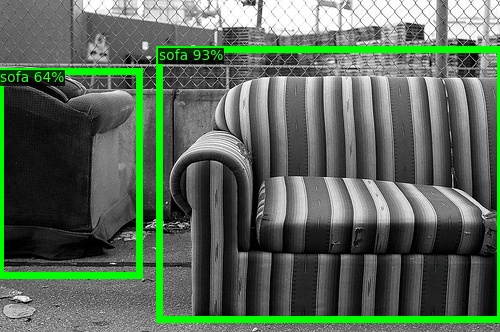} \\
            \rotatebox{90}{\small FSCE \cite{fscontrastive}} &
                \includegraphics[width=0.23\textwidth, height=0.19\textwidth]{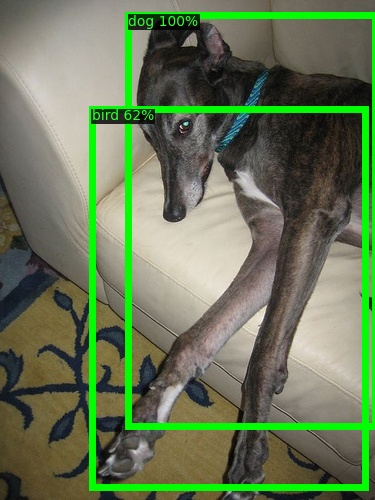} & 
                \includegraphics[width=0.23\textwidth]{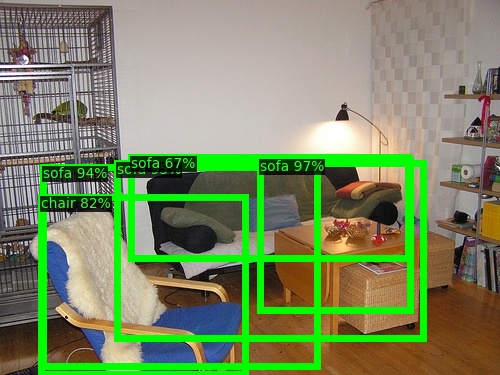} & 
                \includegraphics[width=0.23\textwidth]{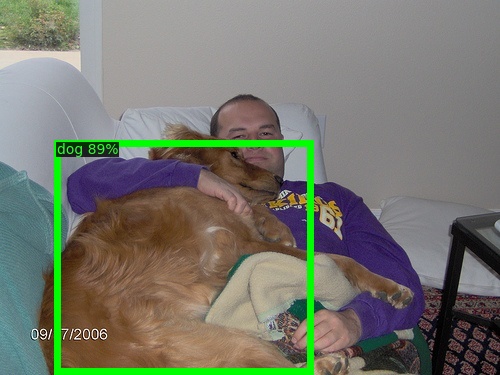} &
                \includegraphics[width=0.23\textwidth]{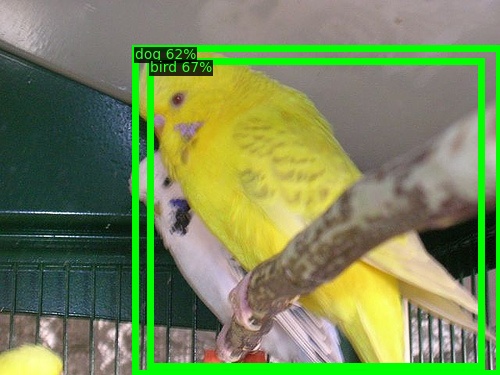} \\
            \rotatebox{90}{\small FSCE + \SMILE\ } & 
                \includegraphics[width=0.23\textwidth, height=0.19\textwidth]{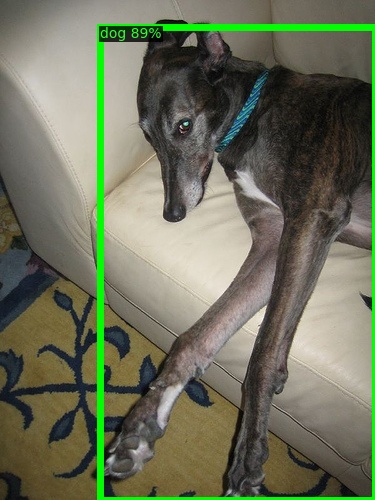} & 
                \includegraphics[width=0.23\textwidth]{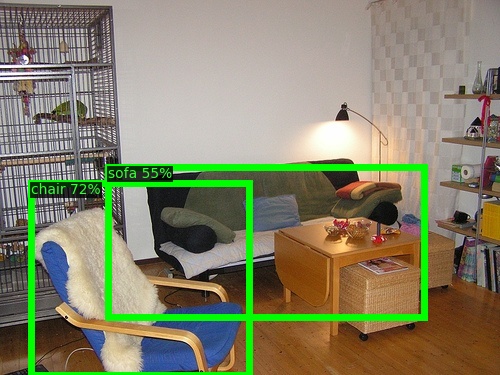} & 
                \includegraphics[width=0.23\textwidth]{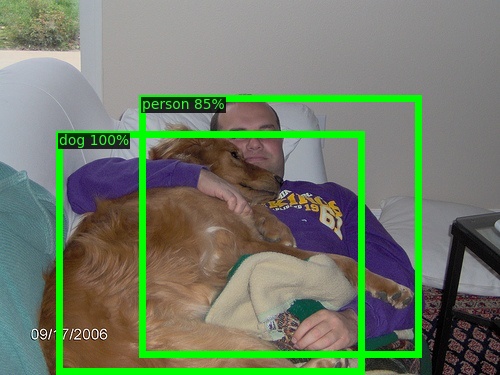} &
                \includegraphics[width=0.23\textwidth]{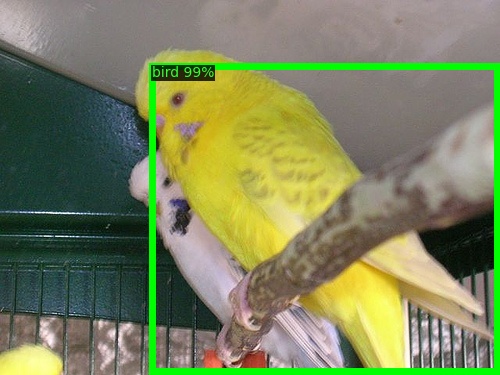} \\
            & (a)  & (b)  & (c) & (d)  \\
        \end{tabular}
        \caption{\textbf{Qualitative results from \SMILE: } We contrast the performance of AGCM and FSCE before and after introduction of the Combinatorial formulation introduced in \SMILE. We observe significant confusion and forgetting in SoTA approaches FSCE and AGCM while introduction of \SMILE\ overcomes most of these pitfalls.}
        \label{fig:qual}
\end{figure*}

%% file: main.bbl
\begin{thebibliography}{10}
\providecommand{\url}[1]{\texttt{#1}}
\providecommand{\urlprefix}{URL }
\providecommand{\doi}[1]{https://doi.org/#1}

\bibitem{agcm}
Agarwal, A., Majee, A., Subramanian, A., Arora, C.: Attention guided cosine margin to overcome class-imbalance in few-shot road object detection. In: Proceedings of the IEEE/CVF Winter Conference on Applications of Computer Vision (WACV) Workshops. pp. 221--230 (2022)

\bibitem{fadi}
Cao, Y., Wang, J., Jin, Y., Wu, T., Chen, K., Liu, Z., Lin, D.: Few-shot object detection via association and discrimination. In: Thirty-Fifth Conference on Neural Information Processing Systems (2021)

\bibitem{lstd}
Chen, H., Wang, Y., Wang, G., Qiao, Y.: {{LSTD:} {A} Low-Shot Transfer Detector For Object Detection}. In: {AAAI}. pp. 2836--2843 (2018)

\bibitem{vit}
Dosovitskiy, A., Beyer, L., Kolesnikov, A., Weissenborn, D., Zhai, X., Unterthiner, T., Dehghani, M., Minderer, M., Heigold, G., Gelly, S., Uszkoreit, J., Houlsby, N.: An image is worth 16x16 words: Transformers for image recognition at scale. In: International Conference on Learning Representations (2021)

\bibitem{voc}
Everingham, M., Van~Gool, L., Williams, C.K.I., Winn, J., Zisserman, A.: {The {Pascal} {Visual} {Object} Classes ({VOC}) {Challenge}}. IJCV pp. 303--338 (2010)

\bibitem{pascalvoc}
Everingham, M., Van~Gool, L., Williams, C., Winn, J., Zisserman, A.: The pascal visual object classes (voc) challenge. International Journal of Computer Vision  \textbf{88},  303--338 (06 2010)

\bibitem{fsod}
Fan, Q., Zhuo, W., Tang, C.K., Tai, Y.W.: {Few-Shot Object Detection With {Attention-RPN} And {Multi-Relation} Detector}. In: {CVPR} (2020)

\bibitem{Fan_2021_CVPR}
Fan, Z., Ma, Y., Li, Z., Sun, J.: {G}eneralized {F}ew-{S}hot {O}bject {D}etection {W}ithout {F}orgetting. In: Proceedings of the IEEE/CVF Conference on Computer Vision and Pattern Recognition (CVPR). pp. 4527--4536 (June 2021)

\bibitem{rrcnn}
Fan, Z., Ma, Y., Li, Z., Sun, J.: Generalized few-shot object detection without forgetting  (2021)

\bibitem{maml}
Finn, C., Abbeel, P., Levine, S.: {{Model-Agnostic} {Meta-Learning} For Fast Adaptation Of Deep Networks}. In: {ICML} (2017)

\bibitem{fujishige}
Fujishige, S.: Submodular Functions and Optimization, vol.~58. Elsevier (2005)

\bibitem{nonforget}
Gidaris, S., Komodakis, N.: {Dynamic Few-Shot Visual Learning Without Forgetting}. In: {CVPR} (2018)

\bibitem{fast-rcnn}
Girshick, R.B.: {Fast {R-CNN}}. ICCV  (2015)

\bibitem{fct}
Han, G., Ma, J., Huang, S., Chen, L., Chang, S.F.: {Few}-shot object detection with fully cross-transformer. In: Proceedings of the IEEE/CVF Conference on Computer Vision and Pattern Recognition. pp. 5321--5330 (2022)

\bibitem{submod_cooperation}
Jegelka, S., Bilmes, J.: Submodularity beyond submodular energies: Coupling edges in graph cuts. In: CVPR 2011 (2011)

\bibitem{reweight}
Kang, B., Liu, Z., Wang, X., Yu, F., Feng, J., Darrell, T.: {Few-shot Object Detection Via Feature Reweighting}. In: {ICCV} (2019)

\bibitem{repmet}
Karlinsky, L., Shtok, J., Harary, S., Schwartz, E., Aides, A., Feris, R., Giryes, R., Bronstein, A.M.: {{RepMet}: Representative-Based Metric Learning For Classification And Few-Shot Object Detection}. In: {CVPR} (2019)

\bibitem{labelverifycorrect}
Kaul, P., Xie, W., Zisserman, A.: {Label, Verify, Correct: A Simple Few-Shot Object Detection Method}. In: IEEE Conference on Computer Vision and Pattern Recognition (2022)

\bibitem{vid_sum_2019}
Kaushal, V., Iyer, R., Doctor, K., Sahoo, A., Dubal, P., Kothawade, S., Mahadev, R., Dargan, K., Ramakrishnan, G.: Demystifying multi-faceted video summarization: Tradeoff between diversity, representation, coverage and importance. In: 2019 IEEE Winter Conference on Applications of Computer Vision (WACV). pp. 452--461 (2019)

\bibitem{supcon2020}
Khosla, P., Teterwak, P., Wang, C., Sarna, A., Tian, Y., Isola, P., Maschinot, A., Liu, C., Krishnan, D.: Supervised contrastive learning. In: Advances in Neural Information Processing Systems (2020)

\bibitem{talisman}
Kothawade, S., Ghosh, S., Shekhar, S., Xiang, Y., Iyer, R.K.: Talisman: Targeted active learning for object detection with rare classes and slices using submodular mutual information. In: Computer Vision - {ECCV} 2022 - 17th European Conference (2022)

\bibitem{prism}
Kothawade, S., Kaushal, V., Ramakrishnan, G., Bilmes, J.A., Iyer, R.K.: {PRISM:} {A} rich class of parameterized submodular information measures for guided data subset selection. In: Thirty-Sixth {AAAI} Conference on Artificial Intelligence, {AAAI}. pp. 10238--10246 (2022)

\bibitem{pdc}
Li, B., Liu, C., Shi, M., Chen, X., Ji, X., Ye, Q.: Proposal distribution calibration for few-shot object detection. IEEE transactions on neural networks and learning systems  (2022)

\bibitem{cme}
Li, B., Yang, B., Liu, C., Liu, F., Ji, R., Ye, Q.: {Beyond Max-Margin: Class Margin Equilibrium For Few-Shot Object Detection}. In: {CVPR} (June 2021)

\bibitem{submod_diversity}
Lin, H., Bilmes, J.: A class of submodular functions for document summarization. In: Proceedings of the 49th Annual Meeting of the Association for Computational Linguistics: Human Language Technologies (2011)

\bibitem{mscoco}
Lin, T.Y., Maire, M., Belongie, S., Hays, J., Perona, P., Ramanan, D., Doll{\'a}r, P., Zitnick, C.L.: Microsoft coco: Common objects in context. In: Computer Vision -- ECCV 2014. pp. 740--755. Springer International Publishing, Cham (2014)

\bibitem{imted}
Liu, F., Zhang, X., Peng, Z., Guo, Z., Wan, F., Ji, X., Ye, Q.: {Integrally Migrating Pre-trained Transformer Encoder-decoders for Visual Object Detection}. In: Proceedings of the IEEE/CVF International Conference on Computer Vision. pp. 6825--6834 (2023)

\bibitem{digeo}
Ma, J., Niu, Y., Xu, J., Huang, S., Han, G., Chang, S.F.: Digeo: Discriminative geometry-aware learning for generalized few-shot object detection. In: Proceedings of the IEEE/CVF Conference on Computer Vision and Pattern Recognition (CVPR) (June 2023)

\bibitem{majee2021fewshot}
Majee, A., Agrawal, K., Subramanian, A.: {Few-Shot Learning For Road Object Detection}. In: AAAI Workshop on Meta-Learning and MetaDL Challenge. vol.~140, pp. 115--126 (2021)

\bibitem{score}
Majee, A., Kothawade, S.N., Killamsetty, K., Iyer, R.K.: {SCoRe}: {S}ubmodular {C}ombinatorial {R}epresentation {L}earning. In: Forty-first International Conference on Machine Learning (ICML) (2024)

\bibitem{faster-rcnn}
Ren, S., He, K., Girshick, R.B., Sun, J.: {{Faster R-CNN:} Towards Real-Time Object Detection With Region Proposal Networks}. IEEE Trans. on Pattern Analysis and Machine Intelligence  (2015)

\bibitem{identification}
Shangguan, Z., Rostami, M.: Identification of novel classes for improving few-shot object detection (2023)

\bibitem{protonet}
Snell, J., Swersky, K., Zemel, R.: {Prototypical Networks For Few-shot Learning}. In: {NeurIPS}. pp. 4077--4087 (2017)

\bibitem{fscontrastive}
Sun, B., Li, B., Cai, S., Yuan, Y., Zhang, C.: {FSCE: Few-Shot Object Detection Via Contrastive Proposal Encoding}. In: {CVPR} (June 2021)

\bibitem{relation-net}
Sung, F., Yang, Y., Zhang, L., Xiang, T., Torr, P.H., Hospedales, T.M.: {Learning To {Compare:} {R}elation Network For Few-Shot Learning}. In: {CVPR} (June 2018)

\bibitem{matching-net}
Vinyals, O., Blundell, C., Lillicrap, T., kavukcuoglu, k., Wierstra, D.: {Matching Networks For One Shot Learning}. In: {NeurIPS} (2016)

\bibitem{fsdet}
Wang, X., Huang, T.E., Darrell, T., Gonzalez, J.E., Yu, F.: {Frustratingly Simple Few-Shot Object Detection}. In: {ICML} (2020)

\bibitem{metadet}
Wang, Y.X., Ramanan, D., Hebert, M.: {Meta-Learning To Detect Rare Objects}. In: {ICCV} (2019)

\bibitem{wu2020mpsr}
Wu, J., Liu, S., Huang, D., Wang, Y.: Multi-scale positive sample refinement for few-shot object detection. In: European Conference on Computer Vision (2020)

\bibitem{addfeat}
Xiao, Y., Marlet, R.: {Few-{S}hot Object Detection And Viewpoint Estimation For Objects In The Wild}. In: {ECCV} (2020)

\bibitem{metarcnn}
Yan, X., Chen, Z., Xu, A., Wang, X., Liang, X., Lin, L.: {{Meta R-CNN:} {T}owards General Solver For Instance-Level Low-Shot Learning}. In: {CVPR}. pp. 9577--9586 (2019)

\bibitem{pnpdet}
Zhang, G., Cui, K., Wu, R., Lu, S., Tian, Y.: {{PNPDet:} {E}fficient {Few-Shot} Detection Without Forgetting Via {Plug-And-Play} Sub-Networks}. In: {WACV}. pp. 3823--3832 (2021)

\bibitem{mutualsupport}
Zhang, L., Zhou, S., Guan, J., Zhang, J.: {Accurate Few-Shot Object Detection With Support-Query Mutual Guidance And Hybrid Loss}. In: {CVPR} (June 2021)

\bibitem{Zhang_2020_ACCV}
Zhang, S., Luo, D., Wang, L., Koniusz, P.: {Few-Shot Object Detection By Second-order Pooling}. In: Proceedings of the Asian Conference on Computer Vision (ACCV) (November 2020)

\bibitem{srr}
Zhu, C., Chen, F., Ahmed, U., Shen, Z., Savvides, M.: {Semantic Relation Reasoning For Shot-Stable Few-Shot Object Detection}. In: {CVPR} (June 2021)

\end{thebibliography}
